\documentclass[journal,twoside,web]{ieeecolor}
\usepackage{jsen}
\usepackage{cite}
\usepackage{amsmath,amssymb,amsfonts}
\usepackage{algorithmic}
\usepackage{graphicx}
\usepackage{textcomp}
\usepackage{wrapfig}
\usepackage{subcaption}
\usepackage{tabularx}
\usepackage{tikz}
\usepackage{multirow}
\usepackage{booktabs} 

\def\BibTeX{{\rm B\kern-.05em{\sc i\kern-.025em b}\kern-.08em
    T\kern-.1667em\lower.7ex\hbox{E}\kern-.125emX}}
\definecolor{abstractbg}{rgb}{0.89804,0.94510,0.83137}
\begin{document}
\title{SRMambaV2: Biomimetic Attention for Sparse Point Cloud Upsampling in Autonomous Driving}
\author{Chuang Chen, Xiaolin Qin, Jing Hu, Wenyi Ge
\thanks{This work was supported by the "Juyuan Xingchuan" Project of Central Universities and Research Institutes in Sichuan—High-Resolution Multi-Wavelength Lidar System and Large-Scale Industry Application(2024ZHCG0190), Sichuan Science and Technology Program (2024NSFJQ0035), the Talents by Sichuan provincial Party Committee Organization Department, Key R\&D Project of the Sichuan Provincial Department of Science and Technology-Research on Three-Dimensional Multi-Resolution Intelligent Map Construction Technology (2024YFG0009), the Intelligent Identification and Assessment for Disaster Scenes: Key Technology Research and Application Demonstration (2025YFNH0008), LiDAR Imaging and Intelligent Data Fusion Processing for Unmanned Platforms (2024NSFTD0044), and Project of the Sichuan Provincial Department of Science and Technology—Application and Demonstration of Intelligent Fusion Processing of Laser Imaging Radar Data (2024ZHCG0176).}
\thanks{Chuang Chen is with the College of Computer Science, Chengdu University of Information
Technology, Chengdu, 610225, China (e-mail: 3230609001@stu.cuit.edu.cn). }
\thanks{S. B. Author, Jr., was with Rice University, Houston, TX 77005 USA. He is 
now with the Department of Physics, Colorado State University, Fort Collins, 
CO 80523 USA (e-mail: author@lamar.colostate.edu).}
\thanks{T. C. Author is with 
the Electrical Engineering Department, University of Colorado, Boulder, CO 
80309 USA, on leave from the National Research Institute for Metals, 
Tsukuba, Japan (e-mail: author@nrim.go.jp).}}

\IEEEtitleabstractindextext{%
\fcolorbox{abstractbg}{abstractbg}{%
\begin{minipage}{\textwidth}%
\begin{wrapfigure}[12]{r}{3in}%
\includegraphics[width=3in]{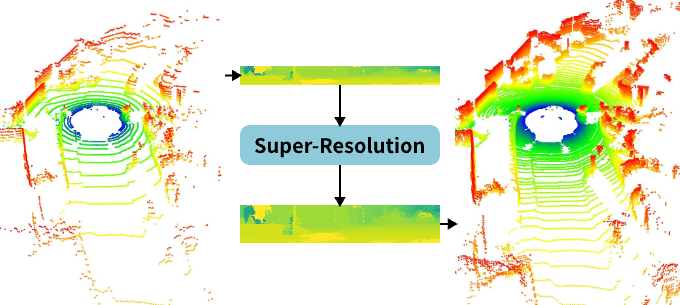}%
\end{wrapfigure}%

\begin{abstract}
Upsampling LiDAR point clouds in autonomous driving scenarios remains a significant challenge due to the inherent sparsity and complex 3D structures of the data. Recent studies have attempted to address this problem by converting the complex 3D spatial scenes into 2D image super-resolution tasks. However, due to the sparse and blurry feature representation of range images, accurately reconstructing detailed and complex spatial topologies remains a major difficulty. To tackle this, we propose a novel sparse point cloud upsampling method named SRMambaV2, which enhances the upsampling accuracy in long-range sparse regions while preserving the overall geometric reconstruction quality. Specifically, inspired by human driver visual perception, we design a biomimetic  2D selective scanning self-attention (2DSSA) mechanism to model the feature distribution in distant sparse areas. Meanwhile, we introduce a dual-branch network architecture to enhance the representation of sparse features. In addition, we introduce a progressive adaptive loss (PAL) function to further refine the reconstruction of fine-grained details during the upsampling process. Experimental results demonstrate that SRMambaV2 achieves superior performance in both qualitative and quantitative evaluations, highlighting its effectiveness and practical value in automotive sparse point cloud upsampling tasks.
\end{abstract}

\end{minipage}}}

\maketitle

\section{Introduction}
\label{sec:introduction}

\IEEEPARstart{L}{ight} detection and ranging (LiDAR) sensors are capable of capturing high-precision three-dimensional structural information at a rate exceeding millions of points per second~\cite{9969624}, demonstrating powerful perceptual capabilities in practical applications such as autonomous driving~\cite{s25092697}, robot navigation~\cite{10657540}, and 3D reconstruction~\cite{11030950}. However, the high hardware cost of LiDAR sensors restricts their widespread deployment in large-scale scenarios. Consequently, a significant amount of research has been devoted to generating dense point clouds from low-cost, sparse point clouds through upsampling, ensuring perceptual accuracy while effectively reducing hardware costs~\cite{10657437,chen2025srmambamambasuperresolutionlidar,10.1007/978-3-031-72784-9_7,SHAN2020103647}.

Range-view-based LiDAR point cloud super-resolution offers a computationally efficient approach to enhancing the geometric details of point clouds by leveraging the structured representation of 2D range images derived from the 3D spatial structure. Several recent methods have focused on learning the multi-scale semantic information within the typical bottom-up architecture~\cite{10657437,SHAN2020103647,FU2025112132}. However, as a projection of 3D point clouds onto a 2D plane, range images exhibit significant variations in information density across different depth regions. This depth-dependent spatial imbalance substantially increases the difficulty for models to perform structural perception in distant areas.

In response to the structural perception challenges caused by this depth-dependent imbalance, we observe that visual attention in the human vision system typically follows a coarse-to-fine processing strategy, where the brain first performs a rough scan to quickly form an overall high-level perception, which is then integrated with sensory input, enabling the brain to make more accurate judgments of the environment~\cite{Saalmann_Pigarev_Vidyasagar_2007,Gilbert_Sigman_2007}. Many previous works have incorporated this coarse-to-fine mechanism as an attention mechanism into image super-resolution tasks~\cite{10.1007/978-3-031-72973-7_21} and other vision models~\cite{10.5555/3524938.3525585,Chen_Sun_Tian_Shen_Huang_Yan_2020}. However,sparse features cause the model to overemphasize strong signals, while critical weak signals are overlooked, leading to insufficient representational power. A key characteristic of the visual attention mechanism is the use of feedback signals as explicit guidance to locate the most important regions in a scene~\cite{Saalmann_Pigarev_Vidyasagar_2007}. However, the classic bottom-up hierarchical architecture adopted by most existing vision backbones extracts signals that lack representativeness~\cite{7780459,7478072}. It progressively encodes features from lower to higher levels while gradually reducing spatial resolution, resulting in high-level features acquiring abstract semantic representations at the cost of losing spatial details.

\begin{figure}
\centering
\begin{subfigure}[b]{0.45\linewidth}
    \includegraphics[width=\linewidth]{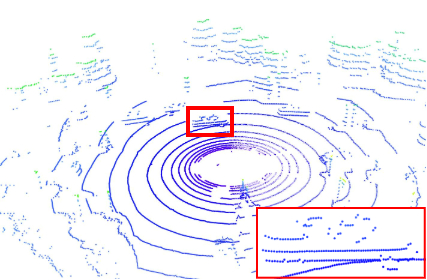}
    \caption{Input}
    \label{fig:input}
\end{subfigure}
\begin{subfigure}[b]{0.45\linewidth}
    \includegraphics[width=\linewidth]{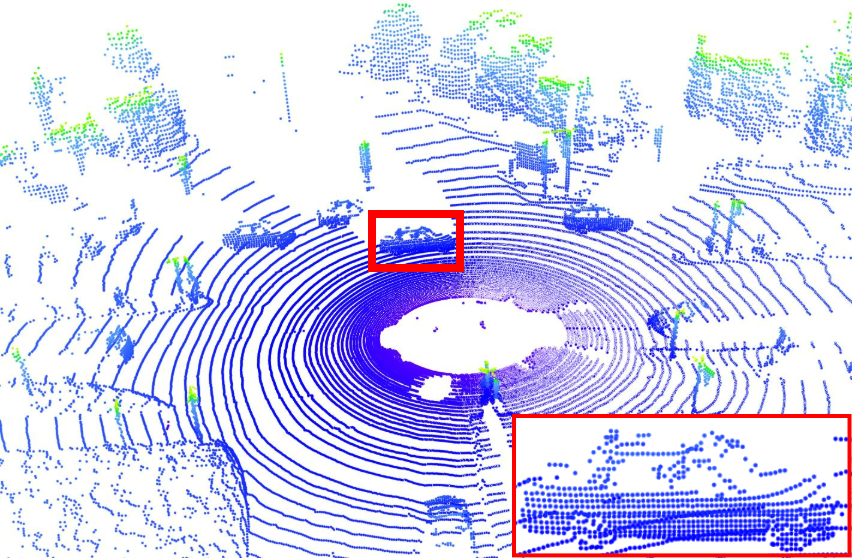}
    \caption{Ground Truth}
    \label{fig:groundtruth}
\end{subfigure}
\begin{subfigure}[b]{0.45\linewidth}
    \includegraphics[width=\linewidth]{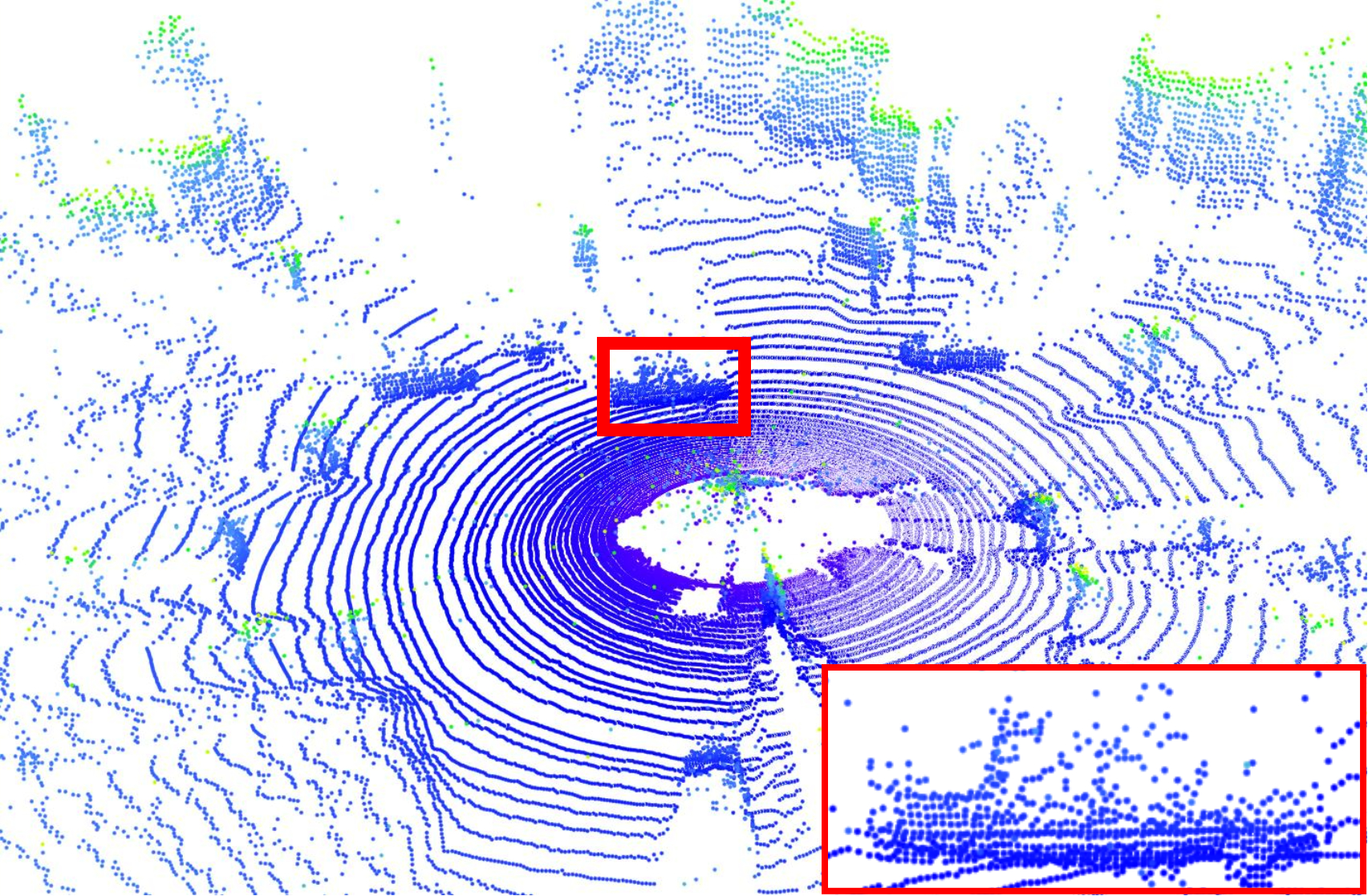}
    \caption{SMFA}
    \label{fig:fbm}
\end{subfigure}
\begin{subfigure}[b]{0.45\linewidth}
    \includegraphics[width=\linewidth]{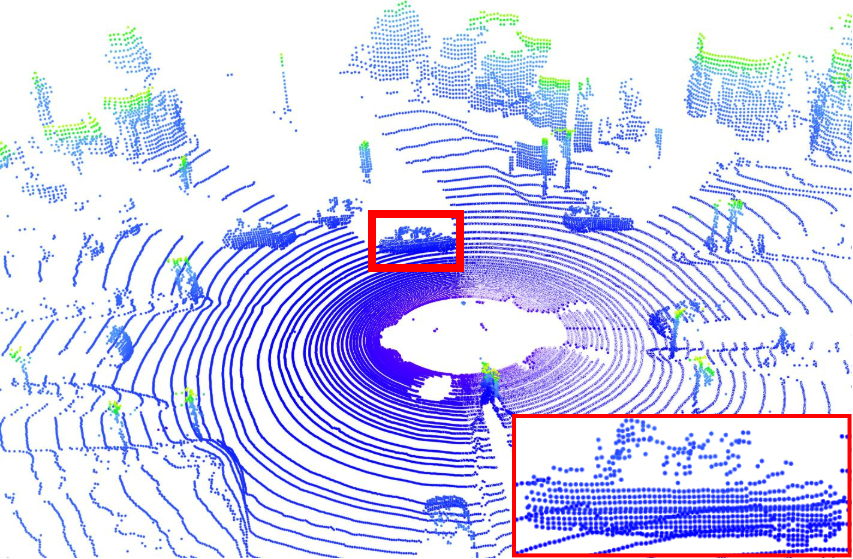}
    \caption{Ours}
    \label{fig:ours}
\end{subfigure}
\caption{Directly applying image super-resolution methods such as SMFA~\cite{10.1007/978-3-031-72973-7_21} to range images introduces significant noise (c), while our approach reconstructs geometrically accurate and visually realistic point clouds (d).}
\label{image-supre-resolution}
\end{figure}

For point cloud super-resolution tasks based on range images, the coarse-to-fine visual attention mechanism is inherently well-suited. Firstly, in autonomous driving scenarios, the visual attention of the driver typically prioritizes key information in distant regions, which aligns with the processing strategy of “coarse perception followed by fine attention”. Secondly, points corresponding to distant areas in range images are usually located in the upper rows of the image, exhibiting a sparse yet relatively concentrated spatial feature distribution. This spatial hierarchy provides effective structural support for extracting and integrating features in a coarse-to-fine, hierarchical manner. 

\begin{figure}
    \centering
    \includegraphics[width=\linewidth]{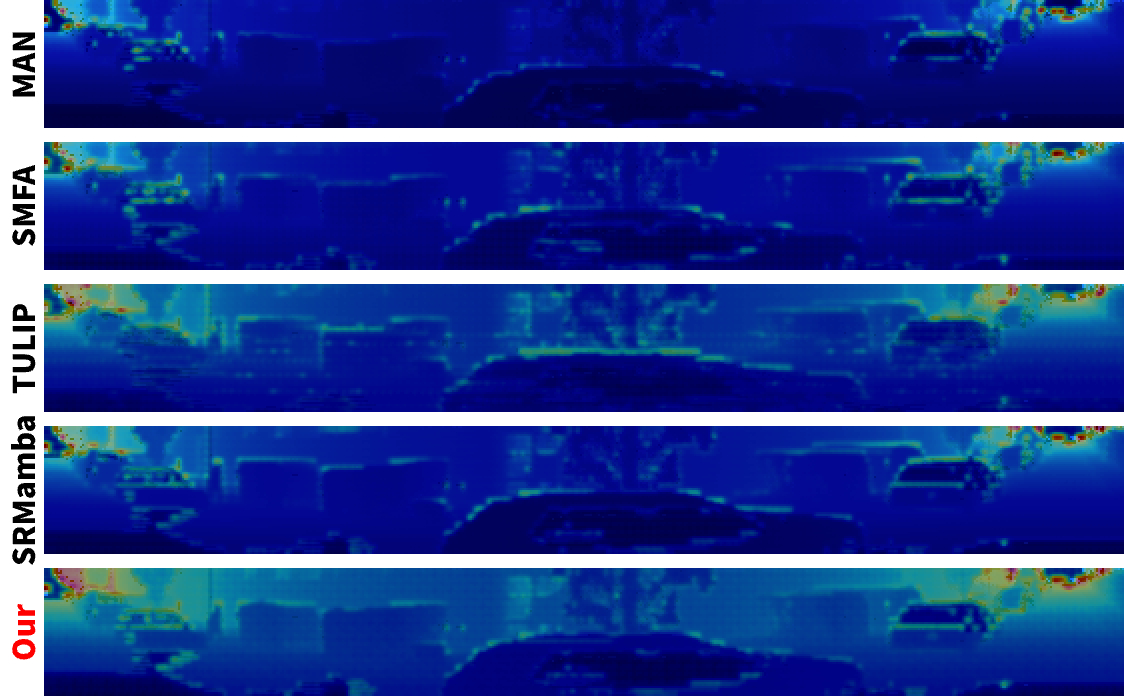}
    \caption{Comparison of heatmap visualization results from different methods on range images, color distributions and activation patterns reflect distinct feature attention characteristics across the approaches. Our proposed visual scan-to-focus mechanism generates features with greater emphasis on global context and demonstrates significantly better performance in sparse feature regions compared to other methods.}
    \label{heatmap_compare}
\end{figure}

To fully extract feedback signals and integrate them into deep feature representations, the “coarse scan” stage should be capable of adaptively modeling long-range dependencies to understand the global structure, to capture fine-grained and essential local details~\cite{Lou_2025_CVPR}. Nonetheless, we find that existing image super-resolution methods cannot effectively adapt to range images, as shown in Fig.~\ref{image-supre-resolution}. Unlike standard RGB images, range images encode depth distribution along the vertical axis while maintaining the continuity of object contours along the horizontal axis. However, attention mechanisms constrained by local windows struggle to capture structures beyond individual windows, exhibiting limited global modeling capability, as shown in Fig.~\ref{heatmap_compare}. We also observe that SRMamb~\cite{chen2025srmambamambasuperresolutionlidar} is capable of effectively extracting object boundary contour information, but it is insensitive to sparse feature regions. Therefore, effectively extracting globally informative feedback signals from range images remains a significant challenge.

Based on the above analysis, we propose a 2D Selective Scanning Self-attention (2DSSA) mechanism inspired by the visual attention strategy of drivers in autonomous driving scenarios. Unlike previous works, we integrate the 2D-Selective-Scan (SS2D)~\cite{liu2024vmamba} mechanism with the attention mechanism~\cite{9710580} to collaboratively enhance the contextual information within feature maps. Specifically, for each input feature map, we extract multi-directional global structural information via an SS2D to generate feedback signals. Subsequently, a semantically informative yet low-fidelity global overview feature map is rapidly constructed. This overview representation is then propagated into the attention-based network to guide the learning of deeper feature representations. In this manner, the proposed biomimetic strategy injects structurally aware global context into each feature layer. Furthermore, we propose a progressive adaptive loss (PAL) function. Recognizing the significant bias of the traditional $L1$ loss in handling distant and sparse regions of point clouds, our loss incorporates a depth-distribution-based weighting scheme and a bird's-eye-view (BEV) consistency constraint. This design enhances reconstruction in distant areas and guides the model to focus on structurally critical regions.

In summary, our main contributions are as follows:
\begin{enumerate}

\item{We propose a 2DSSA architecture, inspired by human visual attention in autonomous driving, which integrates bidirectional scanning with attention to enhance contextual understanding in feature representations.}

\item{We design a structure-aware loss function PAL that leverages depth-based weighting and BEV consistency to enhance reconstruction in distant and structurally significant regions.}

\item{Evaluation experiments demonstrate the proposed method effectively enhances geometric details through upsampling in sparse point cloud scenarios, achieving state-of-the-art performance in both qualitative and quantitative metrics.} 
\end{enumerate}

\section{RELATED WORK}

\subsection{Point cloud super-resolution}

Point cloud super-resolution, as an important task in point cloud processing, aims to convert sparse point cloud data into a dense representation to improve the accuracy of environment perception.

Traditional methods typically rely on interpolation techniques to generate new points, but they struggle to handle the complexities of real-world scenes~\cite{964489,10.1145/1073204.1073227}. In recent years, processing point cloud data using neural networks has become the mainstream approach. Point-based deep learning methods, such as PointNet~\cite{8099499} and its variants~\cite{10.5555/3295222.3295263}, operate directly on raw 3D coordinates.  Other works voxelize point clouds and extract features using 3D convolutions to generate dense points~\cite{9969624,9794916,10944132}. However, their unstructured nature leads to high computational costs.

In LiDAR environment perception, many methods adopt a compact representation of range images~\cite{9969541,9612185}, where range-image-based LiDAR point cloud super-resolution has seen significant progress in recent years. The technique aims to construct dense scenes from sparse observations, often leveraging advances in Convolutional Neural Networks (CNNs) to enhance the fidelity and detail of range images for reduced computational demands. Inspired by the success of image super-resolution methods, Shan et al.~\cite{SHAN2020103647} addressed 3D Euclidean point cloud upsampling by transforming it into a 2D image super-resolution problem, resolved via a deep CNNs. However, traditional CNNs struggle to effectively handle range images with significant variations in scale. To efficiently extract features from range images, TULIP~\cite{10657437} introduces the Swin-Transformer~\cite{9710580} to capture multi-scale information. By modeling long-range dependencies and emphasizing salient features, the attention-based architecture demonstrates superior capability in extracting meaningful representations from range images. However, this design deviates from the human visual mechanism, resulting in the loss of spatial details and contextual information during the early stages of feature extraction, making it difficult to capture the spatial structure of sparse point clouds at long distances.

\subsection{Visual State Space Model}

Visual State Space (VSS) models have been demonstrated to be effective in enhancing feature representation through SS2D modeling, serving as a universal backbone for visual tasks~\cite{liu2024vmamba,10800127,10836868,10962143}. Some methods have also proposed architectures that combine VSS with U-Net~\cite{U-Net}, enabling the fusion of low-level and high-level features~\cite{10958038,10.1007/978-981-97-5609-4_33,10.1007/978-981-96-2882-7_10}. Moreover, to enable efficient computation, some studies introduce local scanning mechanisms~\cite{He_2025_CVPR} and other variants~\cite{10.1007/978-3-031-91979-4_2}, effectively modeling local dependencies. Specially, recent studies redesign hybrid architectures by integrating self-attention modules to efficiently model long-range spatial dependencies, achieving strong performance in downstream vision tasks~\cite{Hatamizadeh_2025_CVPR}. However, prior methods struggle to handle the sparsity inherent in distance image inputs and exhibit limited capacity in capturing relations among compact visual features. While SRMamba~\cite{chen2025srmambamambasuperresolutionlidar} enhances feature input through projection compensation, effectively recovering fine-grained structures, there exists systematic positional bias noise in the upsampling points within distant sparse regions, causing significant deviations between the spatial coordinates of the points and the true target point cloud.

\subsection{Biomimetic Vision Models}

Inspired by the perceptual mechanisms of the human visual system, a large number of visual backbone networks have emerged. For instance, some advanced works, inspired by multi-scale processing mechanisms inherent in the human visual system, have designed multi-scale receptive fields to significantly enhance model performance~\cite{8099589,9711179}. Meanwhile, transformers~\cite{dosovitskiy2020image}, inspired by the mechanism of selective focus in the human visual system, have also been widely applied to various vision-related tasks, such as image super-resolution and even point cloud upsampling. Specifically, MAN~\cite{10678536} employs multi-scale large kernel attention(MLKA) and gated spatial attention unit(GSAU) to aggregate spatial context. TULIP~\cite{10657437} deploys a Swin-Transformer~\cite{9710580} based U-Net architecture to reconstruct sharp details in the range images while reducing noisy points in the prediction. Limited by the computational complexity of Vision Transformers (ViTs) ~\cite{dosovitskiy2020image}, some studies have adopted VSS models to enhance computational efficiency. While these methods exhibit strong performance in the visual domain, their limited capacity to process fine-grained feedback signals for identifying salient regions renders them suboptimal for perceiving sparse range images. OverLocK~\cite{Lou_2025_CVPR} introduces a pure convolutional neural network backbone with a top-down attention to perform finer-grained perception. Inspired by this, we propose a novel attention-based visual network built upon SS2D. Unlike previous works, our method focuses on sparsely distributed edge regions and achieves significant performance improvements in point cloud upsampling tasks.

\section{PROPOSED METHOD}

We seamlessly reformulate the complex and unordered 3D point cloud super-resolution problem into a 2D image super-resolution task. The range image contains spatial distance information rather than RGB visual content, making it fundamentally different from conventional image super-resolution tasks. Consequently, the quality of the range image directly affects the performance of point cloud upsampling. Inspired by SRMamba~\cite{chen2025srmambamambasuperresolutionlidar}, we adopt Hough Voting and Depth Inpainting strategies to enhance the quality of the range image $I$. Due to the resolution disparity of range images (e.g., 8×1024 and 16×1024), where the horizontal direction typically contains richer continuous features, we adopt horizontal one-dimensional convolution to efficiently encode lateral information, generating latent low-dimensional feature $I_{latent}$, with a dimension of $H\times \frac{W}{4} \times C_{1}$. After optimizing the range images, a naive application of state-of-the-art image super-resolution networks to range images still fails to achieve satisfactory performance, as shown in Fig.~\ref{image-supre-resolution}. To tackle this challenge, we propose a novel point cloud upsampling method, SRMambaV2, which adopts a Scan-to-Focus (S2F) strategy to better adapt to range images.

\begin{figure*}
    \centering
    \includegraphics[width=\linewidth]{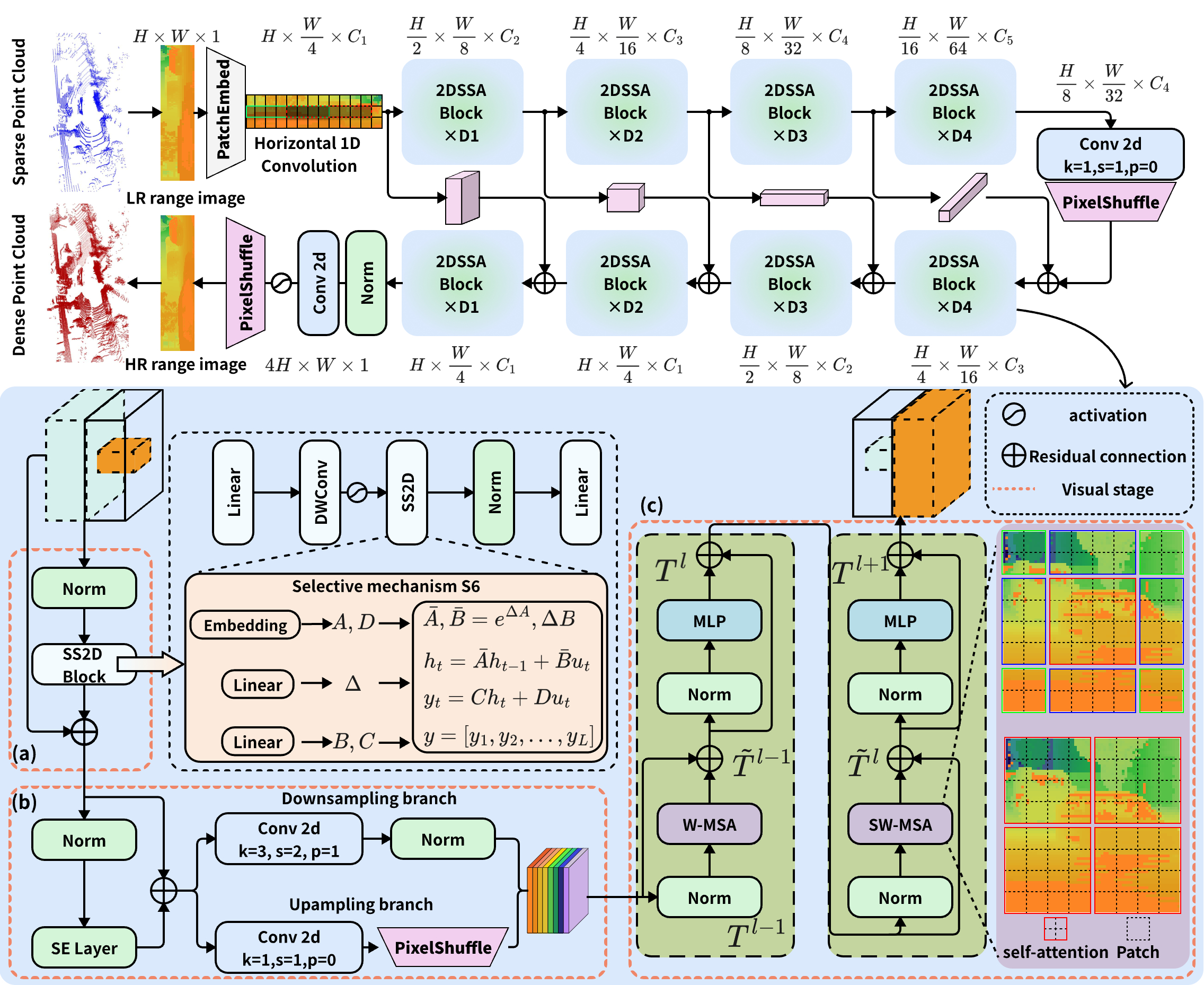}
    \caption{Network architecture of the proposed method. The method uses a U-Net architecture, employing 2DSSA as the encoder and decoder, which comprises three main steps: (a) scanning, (b) modulation, and (c) focus.}
    \label{architecture}
\end{figure*}

\begin{figure}
    \centering
    \includegraphics[width=\linewidth]{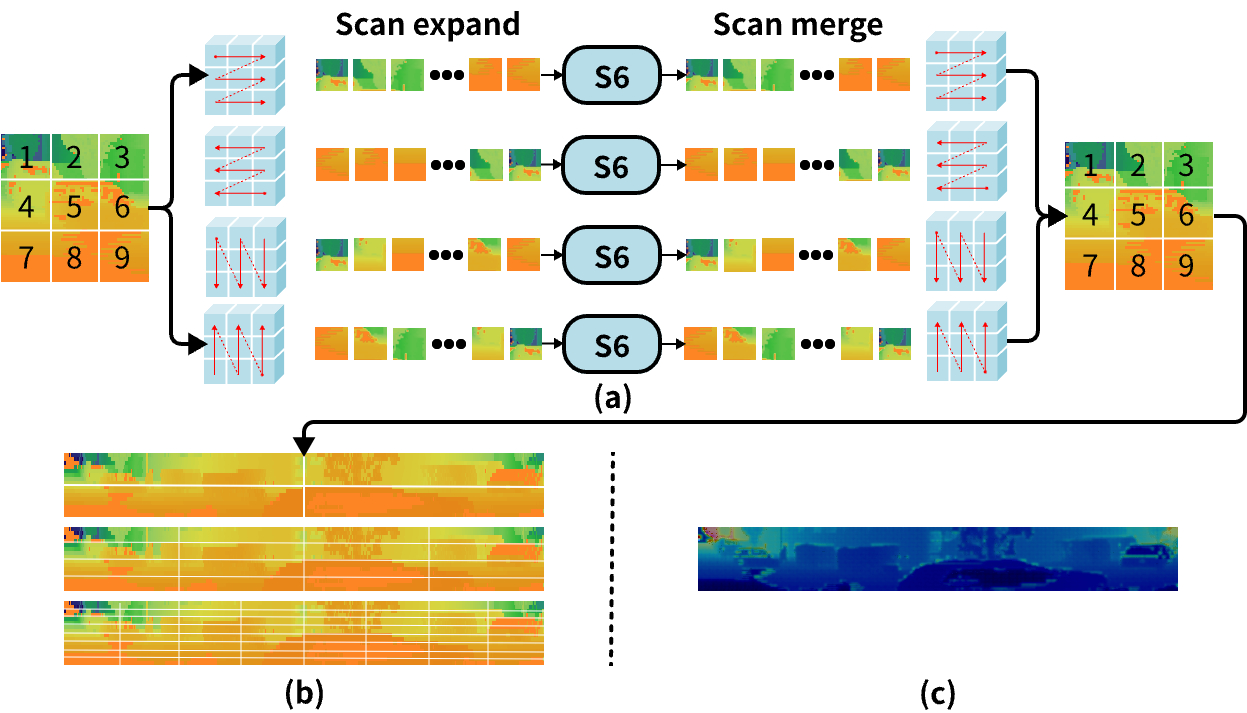}
    \caption{Demonstration of the global feature extraction process using 2DSSA. (a) SS2D performs an initial global feature scan to guide the subsequent attention mechanism. (b) Fine-grained features are further captured through multi-head self-attention modules with both regular window-based (W-MSA) and shifted window-based (SW-MSA). (c) Global key region capture with heatmap visualization. }
    \label{heatmap}
\end{figure}

\subsection{Problem Definition}
Given a sparse point cloud scene composed of $n$ points $P_{LR}=\{p_1,p_2, \ldots,p_n\}$, where each point has attributes representing its 3D spatial coordinates $p_i=\{x_i,y_i,z_i\}$. Vehicular LiDAR systems offer a wide horizontal field of view (FoV) but capture limited information in the vertical direction. We aim to enhance the feature representation of sparse point clouds along the vertical axis, $P_{HR}=\{p_{1},p_{2},...,p_{n},...,p_{4\times n} \}$. Extending to the image super-resolution task, the initial range image resolution is $H\times W$ as a low-resolution (LR) image, which is upsampling to $4H\times W$ as a high-resolution (HR) image, where $H$ represents the number of LiDAR beams, $W$ denotes the horizontal field of view, and $W$ is set to 1024 in all experiments. The pixel location $(u,v)$ of a 3D point is derived through a spherical projection model, where these coordinates correspond to the point’s elevation and azimuth angles in the LiDAR point cloud, combined with coordinate adjustments. Detailed formulas are provided in Appendix A.

\subsection{Model overview}
Our network architecture is built upon a U-shaped structure, utilizing skip connections between the encoder and decoder to facilitate multi-scale feature fusion. Motivated by the observation that drivers tend to focus on distant targets, we introduce a new S2F strategy comprising three key stages: scanning, modulation, and focusing. As shown in Fig.~\ref{architecture}, the scanning stage produces a low-level feature map by employing a SS2D to browse the input image while integrating global contextual information. Due to the blurred contours near the object boundaries in the range image, the low-level feature map is fed into the modulation stage to enhance feature representation. Specifically, the modulation stage consists of two branches designed to align with the U-Net architecture: the downsampling branch in the encoding stage serves to enlarge the receptive field, while the upsampling branch in the decoding stage facilitates the recovery of geometric and spatial details. The feature is fed into the Swin-Transformer as a context prior, leveraging the W-MSA and SW-MSA to progressively capture salient features of key regions across multi-level spaces, enhancing the perception and representation of sparse boundary details.

\subsubsection{scanning}

As shown in Fig.~\ref{architecture}(a), we adopt the SS2D block as the building block of the scanning stage, which consists of a Linear layer, a Normalization layer, a depth-wise convolution (DWConv), and the SS2D module. We denote the input of the i-th block as $F_{in}\in{R^{c\times h \times w}}$, which is first fed into a LayerNorm to normalize the feature distribution. The output is then forwarded to an SS2D block for global overview of the image, as illustrated in Figure~\ref{heatmap}(a). Selective scanning is performed in four directions—left to right, right to left, top to bottom, and bottom to top—to achieve global perception and highlight feature-rich regions. Subsequently, the output is combined with the input of this stage via a residual connection, generating the initial visual scanning features $F_{scan}\in{R^{c\times h \times w}}$, which can be formulated as:

\begin{equation}
    F_{scan} = Residual(SS2D(Norm(F_{in})),F_{in})
\end{equation}

\subsubsection{modulation}
The pipeline of modulation is given in Fig.~\ref{architecture}(b), the modulation stage takes the $F_{scan}$ as its input. Unlike standard RGB images, range images encode only a single modality—depth—without incorporating rich color cues. This inherent limitation of single-modality input hampers the ability to localize salient features, particularly in regions that require fine-grained contextual understanding. The module primarily consists of an SELayer~\cite{Hu_Shen_Sun_2018} and a dual-branch structure, with one branch responsible for downsampling to enlarge the receptive field during the encoding stage, and the other branch responsible for upsampling to restore geometric details during the decoding stage. The branch operation is applied exclusively to the final layer of each stage with depth $D$, and is omitted in the preceding $D-1$ layers. Additionally, we modified the  ELayer~\cite{Hu_Shen_Sun_2018} by replacing global average pooling (AvgPool) with global max pooling (MaxPool) to enhance its adaptability to salient feature regions. For the downsampling branch, we adopt a 3×3 convolution with stride 2 to enlarge the receptive field, and further increase the channel dimension through a Norm layer, as $F_{down}\in R^{2c\times \frac{h}{2} \times \frac{w}{2}}$. For the upsampling branch, a 1×1 convolution is employed to expand the channel dimension by a factor of 2, followed by a PixelShuffle operation to rearrange the channel features into spatial resolution, recovering fine-grained details, as $F_{up}\in R^{\frac{c}{2}\times 2h \times 2w}$ , which can be formulated as:

\begin{equation}
 \left\{\begin{matrix}
 & F_{mid} = Residual(SE(Norm(F_{scan})),F_{scan}) \\
  &F_{down}=Norm(Conv2d(F_{mid}))) \\
 &F_{up}=PixelShuffle(Conv2d(F_{mid}))
\end{matrix}\right.
\end{equation}

\subsubsection{Focus}
We model the human visual focusing process through an attention mechanism. Although ViTs \cite{dosovitskiy2020image} have been widely applied across various image domains, their quadratic computational complexity makes them unsuitable for handling large-scale super-resolution images.

To further refine feature representation and capture long-range dependencies, we adopt the Swin-Transformer~\cite{9710580} as the core component in the focusing stage, as shown in Fig.~\ref{architecture}(c). Unlike traditional convolutional networks limited by local receptive fields, the Swin-Transformer employs W-MSA and SW-MSA mechanisms that not only enable efficient global context modeling but also feature a hierarchical architecture, making it inherently suited for multi-scale image representation and processing. Specifically, the module partitions the input feature map into non-overlapping patches. In our implementation, we adopt a patch size of $2\times8$ to better accommodate the spatial characteristics of range images. Furthermore, we integrate the output of the modulation stage as a context prior to guide the attention mechanism. This allows the Transformer to attend more effectively to salient spatial structures identified in earlier stages, further improving its representation of fine-grained geometric variations.

As shown in Fig.~\ref{heatmap}, the global feature extraction process involves an initial selective scanning to extract low-level features, followed by the application of a shifted window-based multi-head attention mechanism to effectively capture global features.

\subsection{Loss Function}

\begin{figure}[!h]
    \centering
    \includegraphics[width=\linewidth]{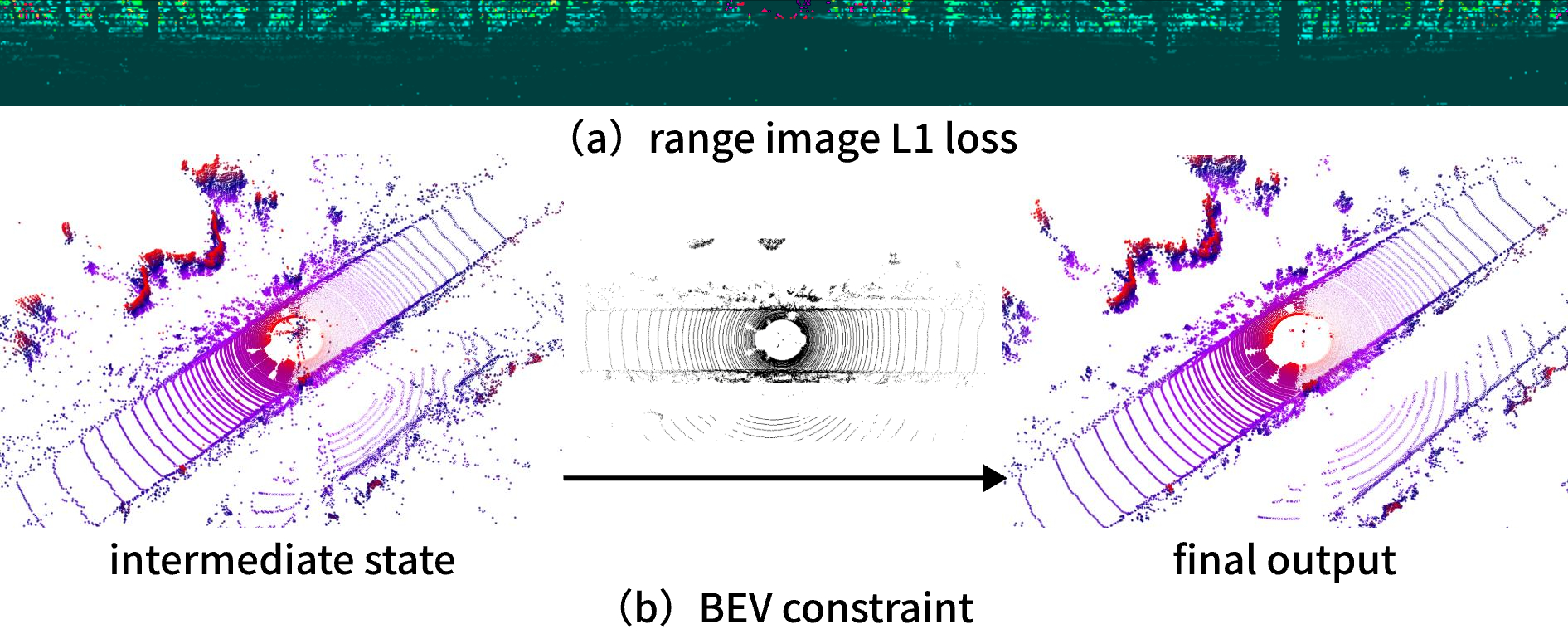}
    \caption{(a) Pixel error in the range image tends to be concentrated in specific regions. (b) BEV constraint effectively mitigates offset in discrete coordinates. }
    \label{loss}
\end{figure}

In most existing super-resolution methods, $L1$ loss is commonly used to optimize pixel-wise accuracy between the reconstructed and ground-truth images. The mathematical expression is given by:

\begin{equation}
    \mathcal{L}_{L1} = \frac{1}{H \times W} \sum_{u=1}^{H} \sum_{v=1}^{W} \left| \hat{I}_{u,v} - I_{u,v} \right|
\end{equation}

where $\hat{I}_{u,v}$ denotes the pixel value at location $(u,v)$ of the reconstructed image, and $I_{u,v}$ denotes the corresponding pixel in the high-resolution ground truth image. $H$ and $W$ represent the height and width of the image, respectively. $L1$ loss demonstrates strong performance in terms of overall pixel-wise error in the upsampled results. However, our objective is not conventional image super-resolution. Relying solely on the $L1$ loss tends to oversmooth high-frequency regions in range images, as shown in Fig.~\ref{loss}(a), which in turn introduces additional noise and geometric artifacts when projecting back to 3D point clouds.

To address this issue, we propose a progressive adaptive loss function comprises three components. First, an $L1$ loss serves as a foundational term to guide the overall convergence of the network. Second, considering the non-uniform feature distribution inherent in range images (e.g., the upper regions of the image, while exhibiting sparse features, often contain concentrated and structurally critical information), we introduce a learnable region-weighted mechanism. This mechanism encourages the network to prioritize these salient regions. Finally, to mitigate 3D structure inconsistencies arising from exclusive reliance on range image supervision, we incorporate a BEV constraint. This constraint provides additional geometric supervision, reducing spatial offsets during point cloud back-projection, as shown in Fig.~\ref{loss}(b). To summarize, our total loss function combines the above three components and is formulated as follows:
\begin{equation}
    \mathcal{L}_{loss} = \alpha\mathcal{L}_{bev} + \beta\mathcal{L}_{adaptiveL1} + \mathcal{L}_{L1}
\end{equation}
The hyperparameters $\alpha$ and $\beta$ control the relative importance of the BEV constraint and the adaptive weighting in the overall loss. Here, $\mathcal{L}_{L1}$ is the standard pixel-wise $L1$ loss that guides the basic reconstruction quality. $\mathcal{L}_{adaptiveL1}$ introduces a learnable weighting mask $M_{u,v}$ to emphasize important regions in the range image, formulated as:
\begin{equation}
 \mathcal{L}_{adaptiveL1}=\frac{1}{H \times W} \sum_{u=1}^{H} \sum_{v=1}^{W} M_{u,v} \left| \hat{I}_{u,v} - I_{u,v} \right|
\end{equation}
Where the mask $M_{u,v}$ is generated via learnable parameters to adaptively weight different spatial regions. The term $\mathcal{L}_{bev}$ represents the BEV constraint, providing geometric supervision in the projected 3D space to reduce structural inconsistencies:
\begin{equation}
    \mathcal{L}_{bev}=\frac{{\textstyle \sum_{x=1}^{H_{bev}}}{\textstyle \sum_{y=1}^{W_{bev}}}|BEV_{pred}(x,y)-BEV_{gt}(x,y)|}{H_{bev}\times W_{bev}} 
\end{equation}
The dimensions $H_{bev}$ and $W_{bev}$ differ from those of the range image, as they are determined by the distribution of point clouds.

\section{Experiment}

\subsection{Experimental Setup}

\subsubsection{Datasets}
To validate the effectiveness of our method, we conduct experiments on two challenging large-scale outdoor datasets: KITTI-360~\cite{liao2022kitti} and nuScenes~\cite{fong2022panoptic}, which contain $360^{^\circ}$ circular area point cloud data captured by Velodyne HDL-64E and HDL-32E sensors, respectively. Building upon previous work, we construct training and validation sets from both datasets. Specifically, for KITTI-360~\cite{liao2022kitti}, we use 20,000 scans for training and 2,500 scans for validation. For nuScenes~\cite{fong2022panoptic}, 28,130 scans are used for training and 6,008 scans for validation.

\subsubsection{Evaluation Metrics}
We use multiple metrics to evaluate the distortion between the original point cloud and the reconstructed point cloud. The upsampling quality of range image directly determines the quality of the back-projected 3D point cloud. We employ the $L1$ loss to evaluate the distortion differences between the upsampled range image and the ground truth. We employ Intersection over Union (IoU), Chamfer Distance (CD), and Jensen-Shannon Divergence (JSD) to comprehensively evaluate point cloud reconstruction quality. IoU assesses voxel-wise spatial overlap, CD measures geometric similarity by nearest neighbor distances, and JSD evaluates distribution consistency between reconstructed and ground truth point clouds.

\subsubsection{Implementation Details}
We downsample the raw point clouds from both datasets by a factor of four to simulate sparse point cloud inputs with different LiDAR beam configurations. The projected range image resolutions are set to 16×1024 and 8×1024, corresponding to different numbers of LiDAR beams, and are upsampled in the vertical direction to 64×1024 and 32×1024, respectively. We train each model for 1000 epochs, saving the model every 50 epochs, and select the model that achieves the best IoU on the validation set.  For optimization, we use AdamW as the default optimizer with an initial learning rate of 0.005. All experiments are conducted on a computer equipped with four NVIDIA Tesla V100-PCIE-16GB, Intel Xeon(R) Silver 4210 CPU (2.20 GHz), and 16 GB of RAM.

\subsection{Benchmark Results}
\subsubsection{Qualitative Evaluation}

We conduct a qualitative comparison of the proposed method against advanced point cloud upsampling approaches, primarily including TULIP~\cite{10657437} and SRMamba~\cite{chen2025srmambamambasuperresolutionlidar}, as well as evaluating the performance of image super-resolution algorithms on point cloud upsampling. All methods are compared using a network depth configuration of [2, 2, 2, 2].

\begin{figure*}
    \centering
    \includegraphics[width=\linewidth]{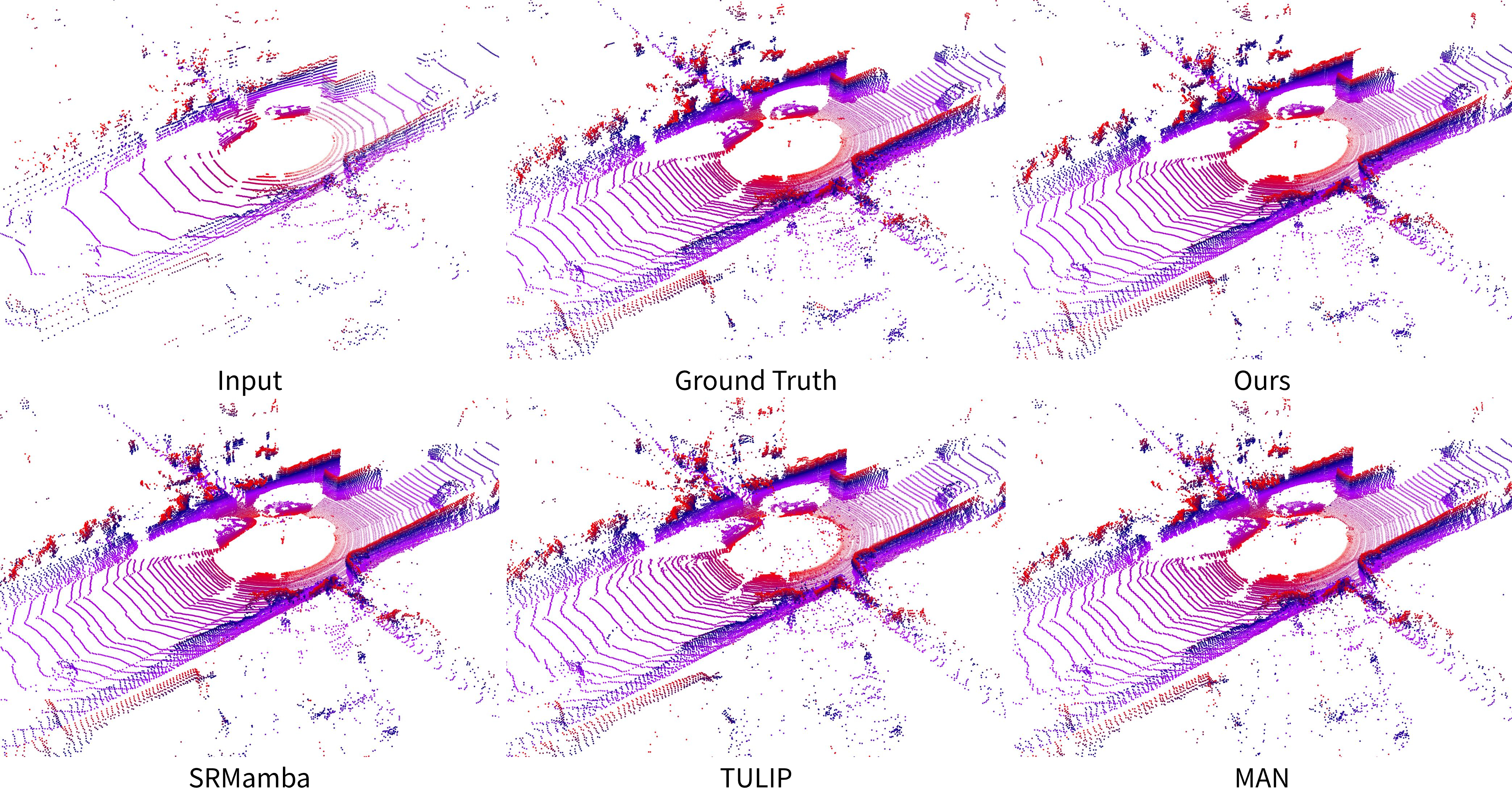}
    \caption{Qualitative visual comparison of the proposed method with other algorithms on the KITTI-360~\cite{liao2022kitti} dataset. The gradient color indicates the height information of the point cloud.}
    \label{KITTI_compare}
\end{figure*}

\begin{figure*}
    \centering
    \includegraphics[width=\linewidth]{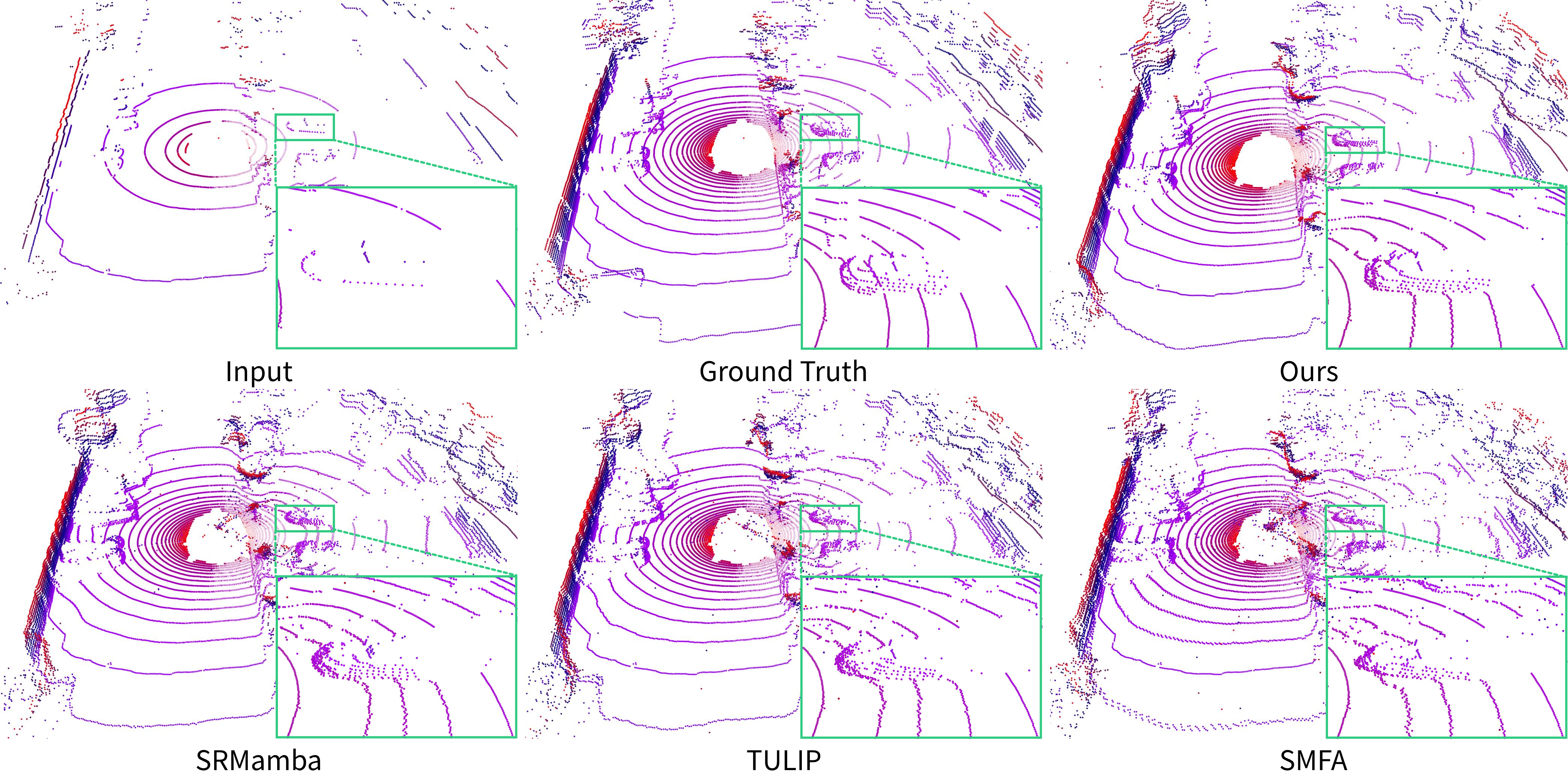}
    \caption{Qualitative visual comparison of the proposed method with other algorithms on the nuScenes~\cite{fong2022panoptic} dataset. For better clarity, key regions of interest are highlighted and magnified using green bounding boxes. }
    \label{nuScenes_compare}
\end{figure*}

Fig.~\ref{KITTI_compare} shows the visualized upsampling results of point clouds for each method. We observe that state-of-the-art image super-resolution methods, despite their advancements, exhibit suboptimal performance when applied to range images. While the MAN~\cite{10678536} approach achieves a MAE comparable to that of TULIP~\cite{10657437}, its performance on three-dimensional evaluation metrics is significantly inferior. TULIP~\cite{10657437} is based on the Swin Transformer and specifically optimized for the characteristics of range images, achieving excellent performance in upsampling tasks, but it still introduces a significant amount of discrete spurious point noise. The VSS-based SRMamba~\cite{chen2025srmambamambasuperresolutionlidar} method demonstrates promising results; however, its limited attention to sparse, long-range regions leads to a gradual degradation in point cloud accuracy with increasing distance. In contrast, our proposed method generates high-quality point clouds that most closely resemble the ground truth. 

\begin{figure}[htb!]
    \centering
    \begin{tabularx}{\columnwidth}{m{0.2cm} @{\quad} >{\centering\arraybackslash}m{\dimexpr\columnwidth-1cm\relax}}
        \rotatebox{90}{Input} & \includegraphics[width=\linewidth]{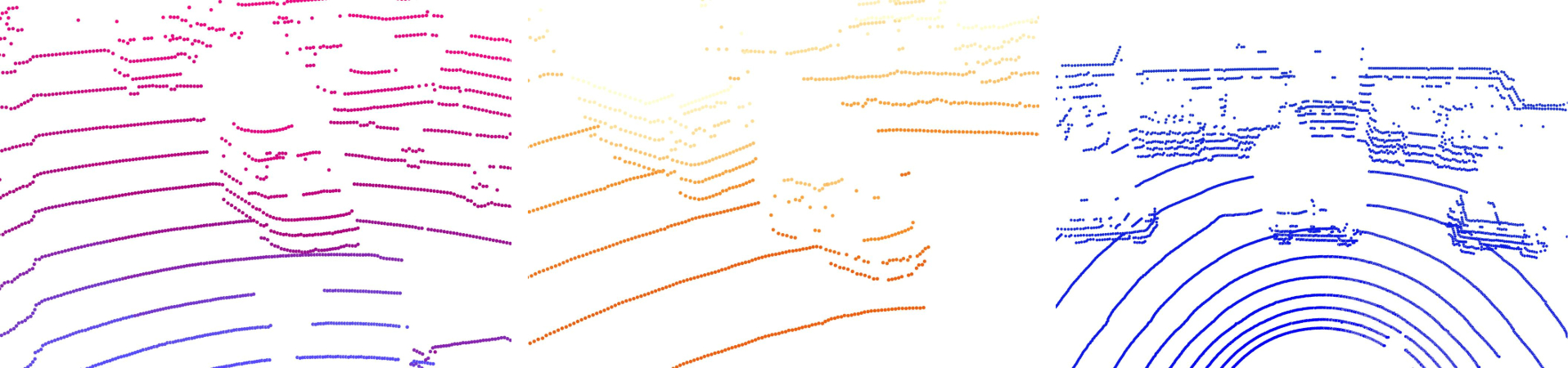} \\

        \rotatebox{90}{Cas-ViT~\cite{zhang2024cas}} & \includegraphics[width=\linewidth]{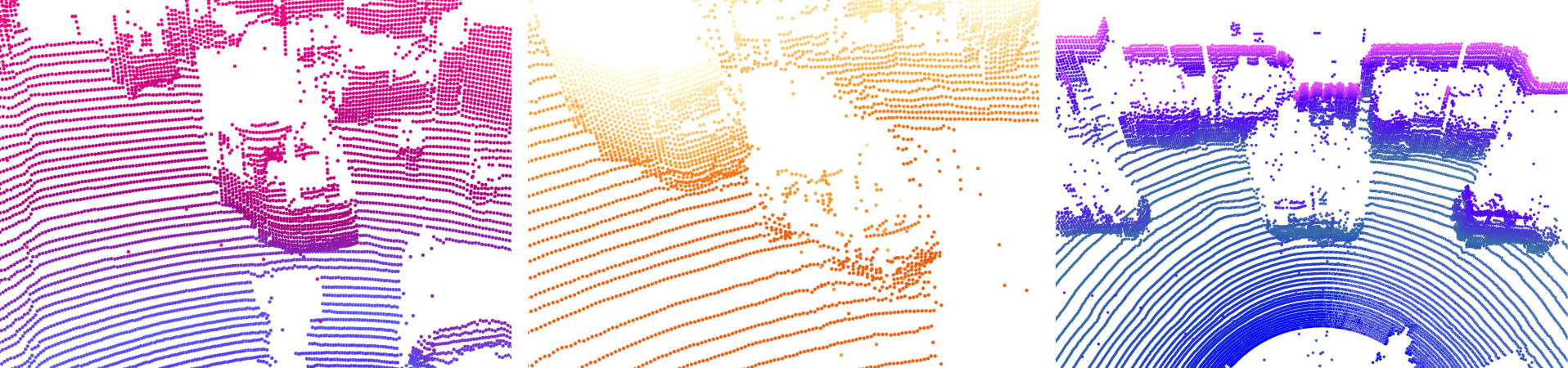} \\

        \rotatebox{90}{SMFA~\cite{10.1007/978-3-031-72973-7_21}} & \includegraphics[width=\linewidth]{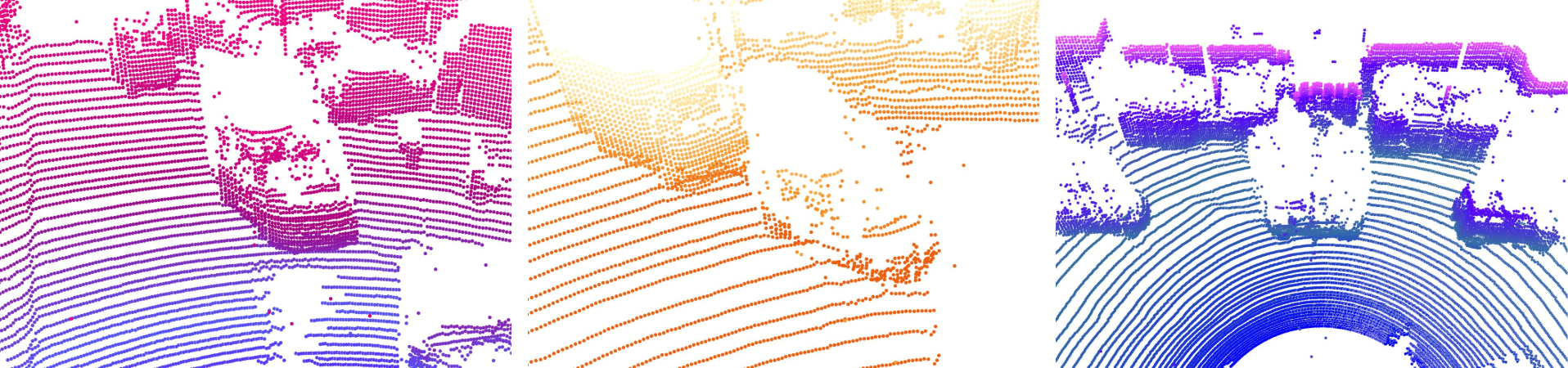} \\

        \rotatebox{90}{MAN~\cite{10678536}} & \includegraphics[width=\linewidth]{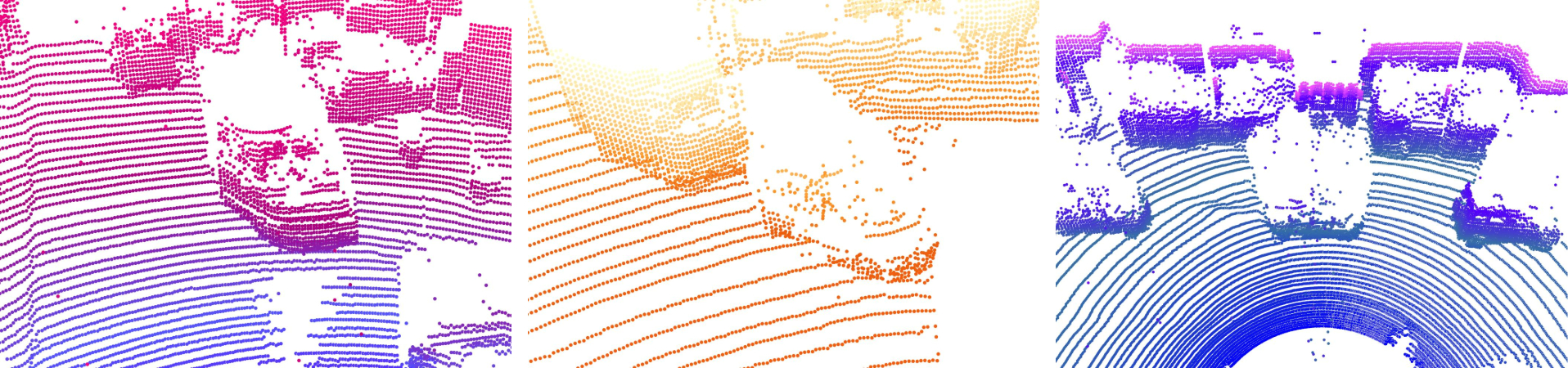} \\

        \rotatebox{90}{TULIP~\cite{10657437}} & \includegraphics[width=\linewidth]{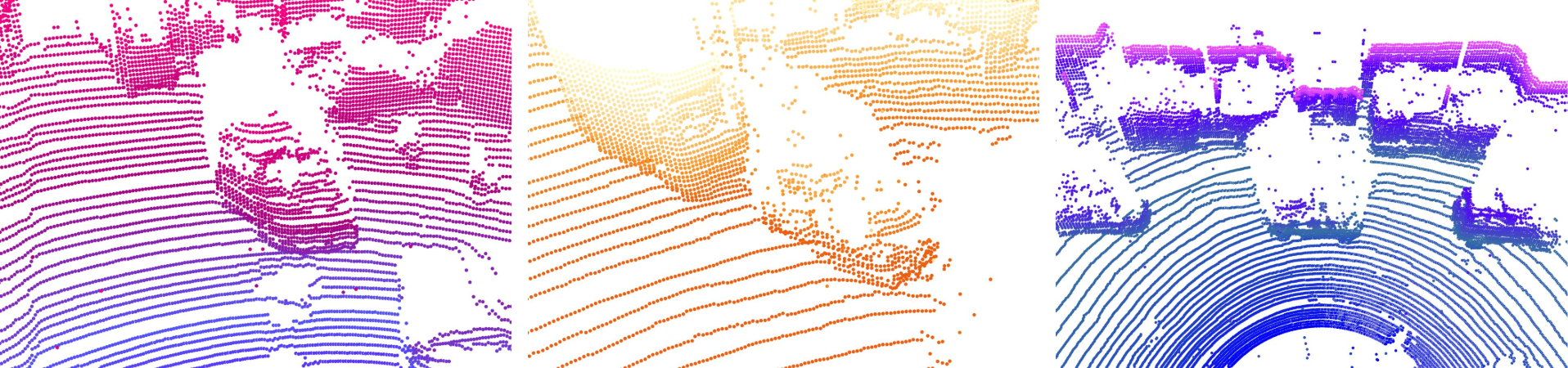} \\

        \rotatebox{90}{SRMamba~\cite{chen2025srmambamambasuperresolutionlidar}} & \includegraphics[width=\linewidth]{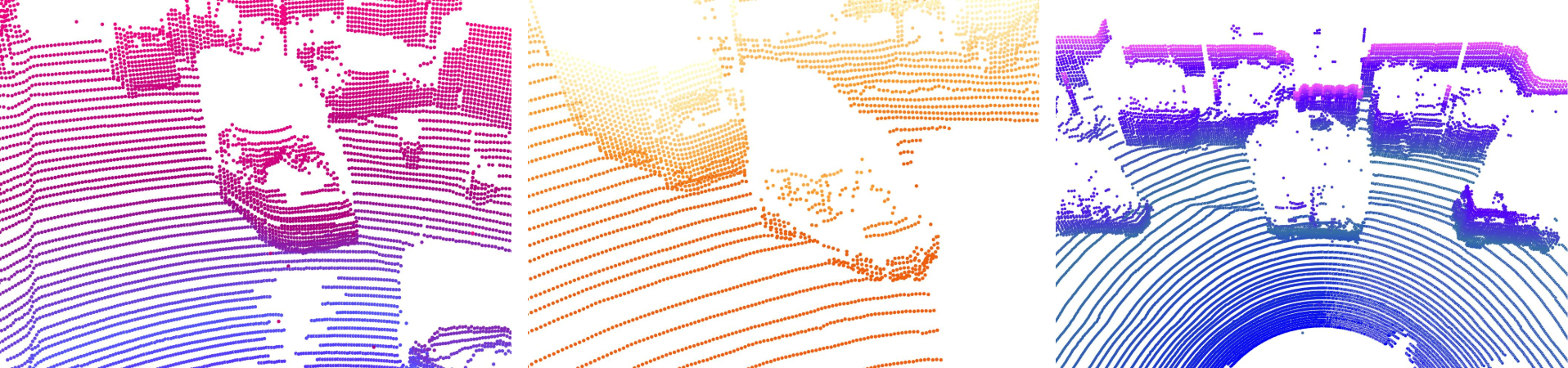} \\

        \rotatebox{90}{Ours} & \includegraphics[width=\linewidth]{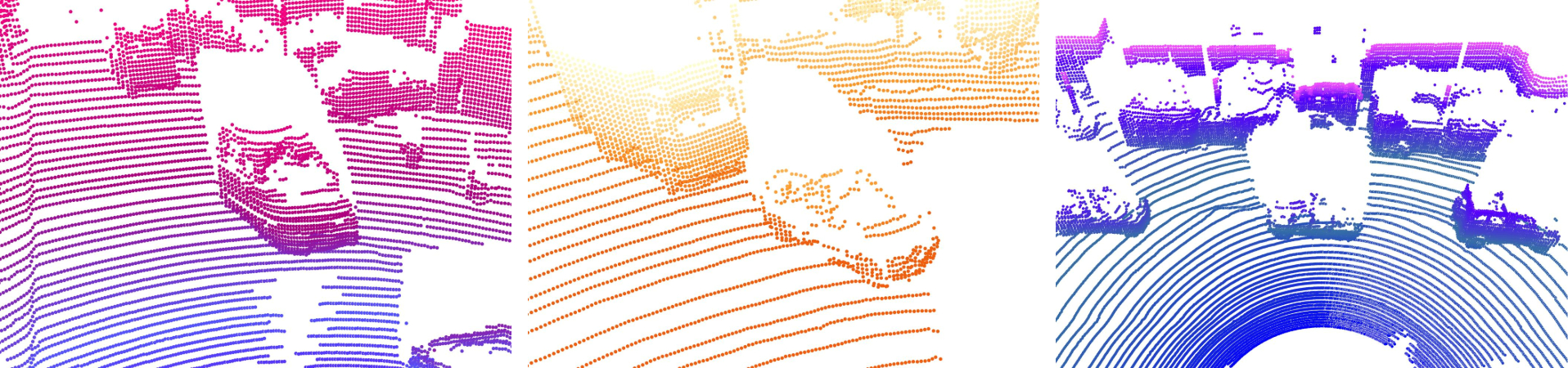} \\

        \rotatebox{90}{GT} & \includegraphics[width=\linewidth]{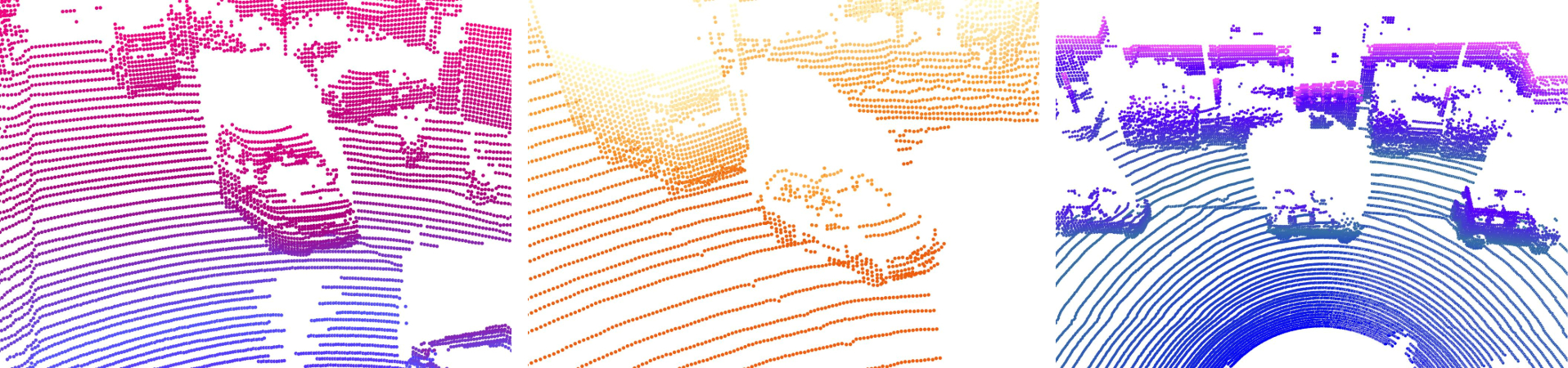} \\
        ~ & \small (a) \quad \quad \quad \quad \quad \quad \quad (b) \quad \quad \quad \quad \quad \quad \quad (c) \\
    \end{tabularx}
    
    \caption{
        Qualitative comparison on KITTI-360~\cite{liao2022kitti}. Our method performs exceptionally well in point cloud upsampling tasks, showing outstanding results particularly in (a) dense car structures, (b) sparse car structures, and (c) complex background interference.
    }
    \label{KITTI_detail_comparsion}
\end{figure}

Fig.~\ref{KITTI_detail_comparsion} provides further illustration of the disparities at the detail level. Experiments show our method surpasses other SOTA approaches in point cloud upsampling, achieving superior reconstruction. (a) In densely populated scenarios, Cas-Vit~\cite{zhang2024cas}, MAN~\cite{10678536}, and SMFA~\cite{10.1007/978-3-031-72973-7_21} reconstruction results exhibit substantial blurring artifacts. TULIP~\cite{10657437} fails to preserve the linear characteristics of LiDAR scans, leading to discontinuous ground contours and fragmented lines, especially in sparse vehicle regions. Conversely, SRMamba~\cite{chen2025srmambamambasuperresolutionlidar} and our method achieve accurate reconstruction of vehicle geometries. (b) Except for SRMamba~\cite{chen2025srmambamambasuperresolutionlidar} and our method, other approaches fail to effectively reconstruct sparse car structures and introduce additional spurious noise points in empty regions. Although SRMamba~\cite{chen2025srmambamambasuperresolutionlidar} produces clear point clouds, the reconstructed vehicle contours exhibit unnatural right-angled bending artifacts. (c) Our method is the only approach capable of avoiding the generation of invalid points under complex background interference while accurately reconstructing the contours of multiple vehicles.

To demonstrate the superiority of our method in sparse scenarios, we conduct visual comparison experiments on selected sparse scenes from the nuScenes~\cite{fong2022panoptic} dataset. According to the qualitative results, as shown in Fig.~\ref{nuScenes_compare} , we can observe the following: under the condition of scanning the car with only two laser beams, SMFA~\cite{10.1007/978-3-031-72973-7_21} and TULIP~\cite{10657437} are limited to reconstructing only a portion of the front region and fail to recover the complete contour structure. SRMamba~\cite{chen2025srmambamambasuperresolutionlidar} is capable of reconstructing the basic contour structure but introduces spurious point noise in the surrounding area. In contrast, the proposed method not only reconstructs the complete contour structure of the car but also effectively suppresses the introduction of artifact noise.

\subsubsection{Quantitative Evaluation}
Quantitative Results on KITTI-360~\cite{liao2022kitti} and nuScenes~\cite{fong2022panoptic}. Table~\ref{Quantitative} presents a quantitative comparison that validates the effectiveness of our proposed SRMambaV2 model. The model demonstrates a clear advantage over all baselines across every metric. In particular, it achieves an order-of-magnitude reduction in the CD, which underscores its remarkable expressive power for generating high-fidelity LiDAR point clouds. 

\begin{table}[h]
\centering
\caption{Comparison of metrics on the KITTI-360 \cite{liao2022kitti} and  nuScenes \cite{fong2022panoptic} Datasets. SRMamba-T denotes shallower network depth, SRMamba-L indicates deeper network. The best-performing results are highlight in bold.}
\begin{tabular}{cccccc}
\hline
Methods & Dataset & IoU $\uparrow$  & CD $\downarrow$  & MAE$\downarrow$ &JSD $\downarrow$ \\
\hline
Bilinear & KITTI-360 & 0.1596 & 1.3110 & 0.0754 & - \\
SMFA\cite{10.1007/978-3-031-72973-7_21} & KITTI-360 & 0.3827 & 0.1577 & 0.0063 & 0.0093 \\
Cas-ViT~\cite{zhang2024cas} & KITTI-360 & 0.3936 & 0.1483 & 0.0076& 0.0091\\
MAN~\cite{10678536} & KITTI-360 & 0.4009 & 1.1258 & 0.0055& 0.0091 \\
Swin-IR\cite{9607618} & KITTI-360 & 0.4077 & 0.1514 & 0.0078  &0.0087 \\
TULIP\cite{10657437} & KITTI-360 & 0.4152 & 0.1241 & 0.0051& 0.0070 \\

SRMamba~\cite{chen2025srmambamambasuperresolutionlidar} & KITTI-360 & 0.4389 & 0.1031 & 0.0044 & 0.0052\\
Ours & KITTI-360 & $\mathbf{0.4516}$ & $\mathbf{0.0826}$ & $\mathbf{0.0041}$ & $\mathbf{0.0045}$\\
\hline
Bilinear & nuScenes & 0.1290 & 2.4178 & 0.0925 & - \\
MAN\cite{10678536} & nuScenes & 0.2760 & 1.1784 & 0.0318 & 0.0380\\
Cas-ViT~\cite{zhang2024cas} & nuScenes & 0.2872 & 1.1624 & 0.0319 & 0.0311 \\
Swin-IR\cite{9607618} & nuScenes & 0.2882 & 1.2527 & 0.0300 &0.0310\\
SMFA\cite{10.1007/978-3-031-72973-7_21} & nuScenes & 0.2932 & 1.1125 & 0.0315  & 0.0308\\
TULIP\cite{10657437} & nuScenes & 0.3048 & 1.0502 & 0.0293  & 0.0304\\

SRMamba~\cite{chen2025srmambamambasuperresolutionlidar} & nuScenes & 0.3170 & 1.0196 & 0.0287  & 0.0293\\

Ours & nuScenes & $\mathbf{0.3299}$ & $\mathbf{0.9485}$ & $\mathbf{0.0284}$ & $\mathbf{0.0267}$\\

\hline
\end{tabular}
\label{Comparison of Metrics on the KITTI Dataset}

\label{Quantitative}
\end{table}

Concretely, our model significantly outperforms the strongest published baseline TULIP~\cite{10657437} across all evaluation metrics. On the KITTI-360~\cite{liao2022kitti} dataset, our approach achieves a remarkable $8.7\%$ absolute gain in IoU, while dramatically reducing CD by $33.4\%$, MAE by $19.6\%$, and JSD by an impressive $35.7\%$. On the more challenging nuScenes~\cite{fong2022panoptic} dataset, our method improves IoU by $8.2\%$, reduces CD by $9.7\%$, MAE by $3.1\%$, and JSD by $12.2\%$. We also compare against SRMamba~\cite{chen2025srmambamambasuperresolutionlidar}, a recent method, where our model achieves further gains: on KITTI-360~\cite{liao2022kitti}, a $2.9\%$ IoU improvement and a $19.9\%$ CD reduction; on nuScenes~\cite{fong2022panoptic}, a $4.1\%$ IoU improvement and a $7.0\%$ CD reduction.

\begin{figure}[htbp]
    \centering
    % 第一行左图
    \begin{minipage}[b]{0.48\columnwidth}
        \centering
        \begin{minipage}[c]{0.05\linewidth}
            \centering
            \rotatebox{90}{\scriptsize IoU}
        \end{minipage}%
        \begin{minipage}[c]{0.95\linewidth}
            \centering
            \includegraphics[width=\linewidth]{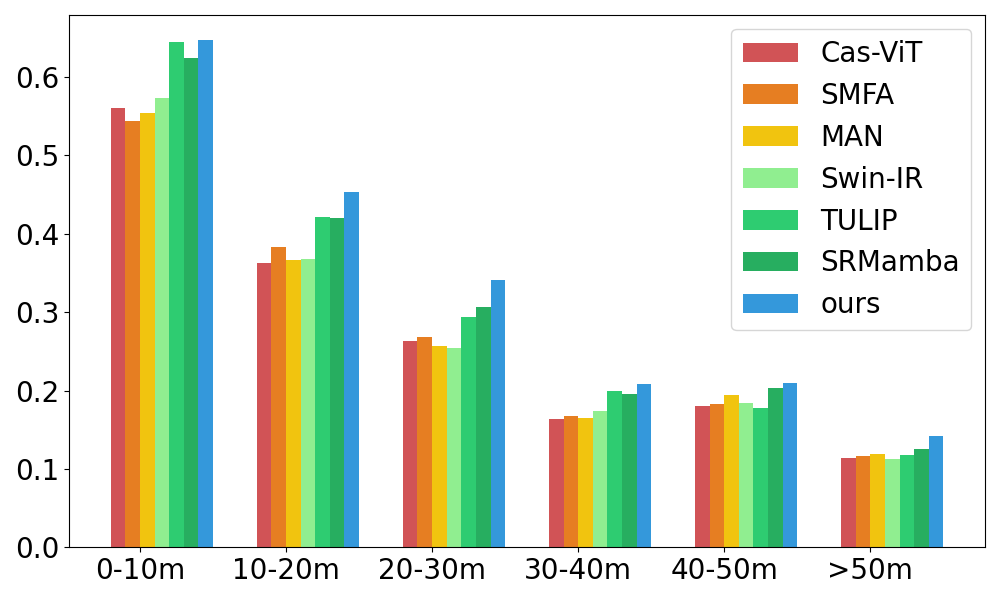} \\
            \scriptsize KITTI-360 IoU $\uparrow$
        \end{minipage}
    \end{minipage}%
    \hfill
    % 第一行右图
    \begin{minipage}[b]{0.48\columnwidth}
        \centering
        \begin{minipage}[c]{0.05\linewidth}
            \centering
            \rotatebox{90}{\scriptsize CD}
        \end{minipage}%
        \begin{minipage}[c]{0.95\linewidth}
            \centering
            \includegraphics[width=\linewidth]{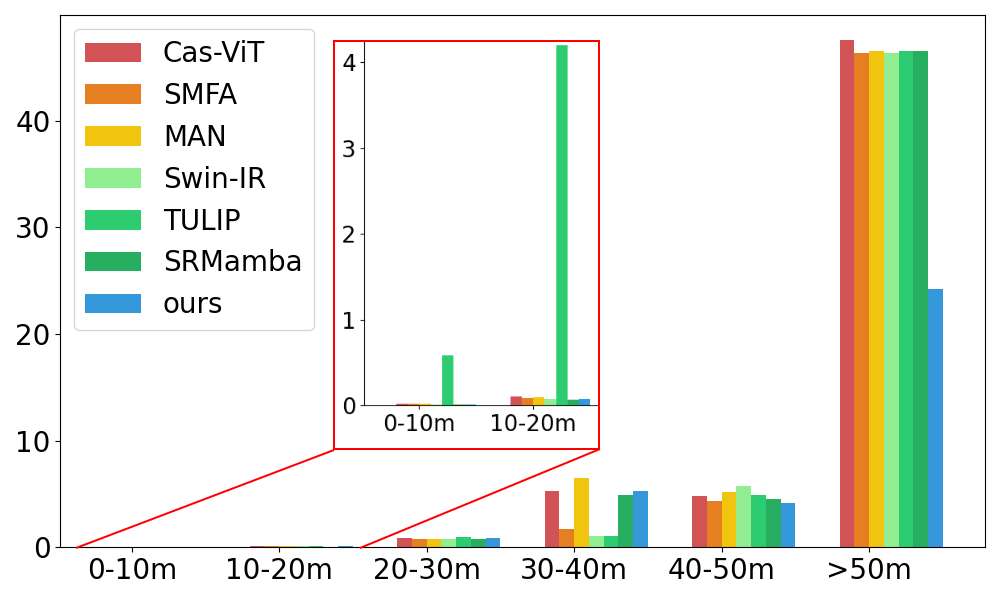} \\
            \scriptsize KITTI-360 CD $\downarrow$
        \end{minipage}
    \end{minipage}

    \vspace{0.3cm}

    % 第二行左图
    \begin{minipage}[b]{0.48\columnwidth}
        \centering
        \begin{minipage}[c]{0.05\linewidth}
            \centering
            \rotatebox{90}{\scriptsize IoU}
        \end{minipage}%
        \begin{minipage}[c]{0.95\linewidth}
            \centering
            \includegraphics[width=\linewidth]{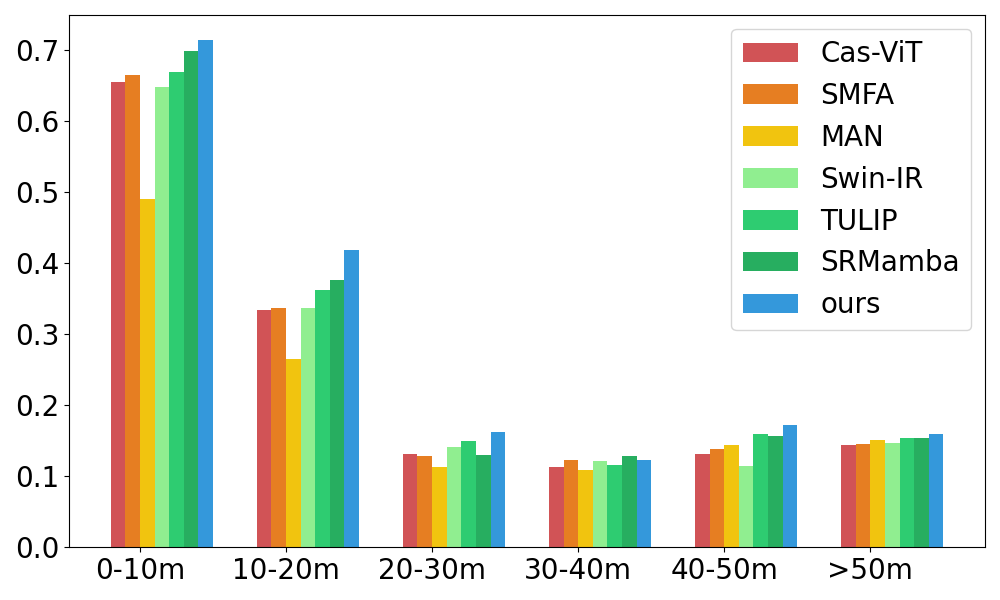} \\
            \scriptsize nuScence IoU $\uparrow$
        \end{minipage}
    \end{minipage}%
    \hfill
    % 第二行右图
    \begin{minipage}[b]{0.48\columnwidth}
        \centering
        \begin{minipage}[c]{0.05\linewidth}
            \centering
            \rotatebox{90}{\scriptsize CD}
        \end{minipage}%
        \begin{minipage}[c]{0.95\linewidth}
            \centering
            \includegraphics[width=\linewidth]{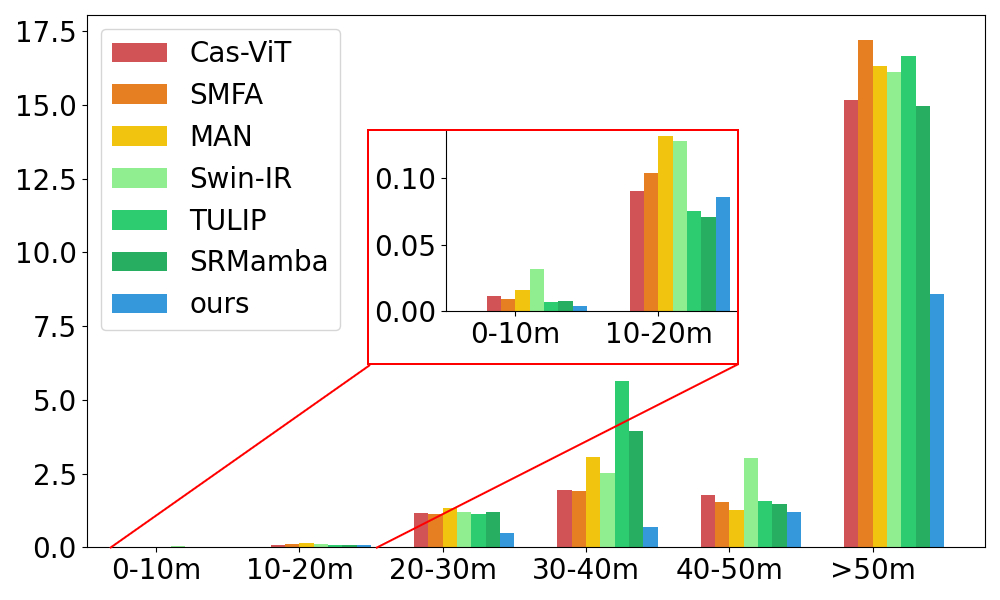} \\
            \scriptsize nuScence CD $\downarrow$
        \end{minipage}
    \end{minipage}

    \caption{IoU and CD metrics visualized across different distance ranges on the KITTI-360~\cite{liao2022kitti} and nuScenes~\cite{fong2022panoptic} datasets.}
    \label{Quantitative_range}
\end{figure}

A fundamental challenge in LiDAR point cloud processing is the inherent degradation of data quality with increasing distance. This sparsity, attributable to sensor FOV limitations and occlusions, poses a significant hurdle for upsampling algorithms as crucial geometric information is progressively lost. To ascertain the robustness of our method against this distance-induced sparsity, we conducted a depth-stratified analysis. As illustrated in Fig.~\ref{Quantitative_range}, we partitioned the point cloud into multiple depth intervals and evaluated the region-wise IoU and CD. The results reveal that our model not only consistently outperforms existing approaches across all depth ranges but also demonstrates a particularly striking improvement in CD. This provides compelling evidence that our approach is adept at preserving salient geometric features even in sparsely sampled regions, mitigating the prevalent artifacts of over-smoothing and detail loss that affect other methods.

\subsection{Ablation Studies}

As demonstrated in Table~\ref{ablation study}, we conduct an ablation study to analyze the contribution of each key component in our framework. By applying 2DSSA to the training, the model dynamically focuses on informative spatial regions, enhancing its ability to capture fine-grained geometric structures. Building upon this, the incorporation of the modulation layer facilitates effective channel-wise feature recalibration, allowing the network to adaptively emphasize geometry-relevant information across channels. Furthermore, the progressive adaptive loss guides the optimization process more effectively by progressively refining the prediction quality, particularly in sparse and geometrically challenging areas.

\begin{table}[!h]
\centering
\caption{Ablation study on KITTI-360 and nuScenes datasets. BL: Base Line. Mod: Modulation. 2DSSA: 2D Selective Scan Attention. PAL: Progressive Adaptive Loss}
\label{tab:ablation}
\resizebox{\columnwidth}{!}{
\begin{tabular}{cccc|ccc|ccc}
\toprule
\multirow{2}{*}{\textbf{BL}} & \multirow{2}{*}{\textbf{2DSSA}} & \multirow{2}{*}{\textbf{Mod}} & \multirow{2}{*}{\textbf{PAL}} & 
\multicolumn{3}{c|}{\textbf{KITTI-360}} & 
\multicolumn{3}{c}{\textbf{nuScenes}} \\
& & & & \textbf{IoU} & \textbf{CD} & \textbf{MAE} & \textbf{IoU} & \textbf{CD} & \textbf{MAE} \\

\midrule
\checkmark & & & & 0.4389 & 0.1031 & 0.0044 & 0.3170 & 1.0196 & 0.0287 \\
\checkmark & \checkmark & & & 0.4416 & 0.0869 & 0.0043 & 0.3199 & 0.9616 & 0.0294 \\
\checkmark & \checkmark & \checkmark & & 0.4468 & 0.0864 & 0.0044 & 0.3210 & 0.9718 & 0.0293 \\
\checkmark & \checkmark & \checkmark & \checkmark & 0.4516 & 0.0826 & 0.0041 & 0.3299 & 0.9485 & 0.0284 \\
\bottomrule
\end{tabular}
}
\label{ablation study}
\end{table}

\begin{figure}[!h]
    \centering
    \includegraphics[width=\linewidth]{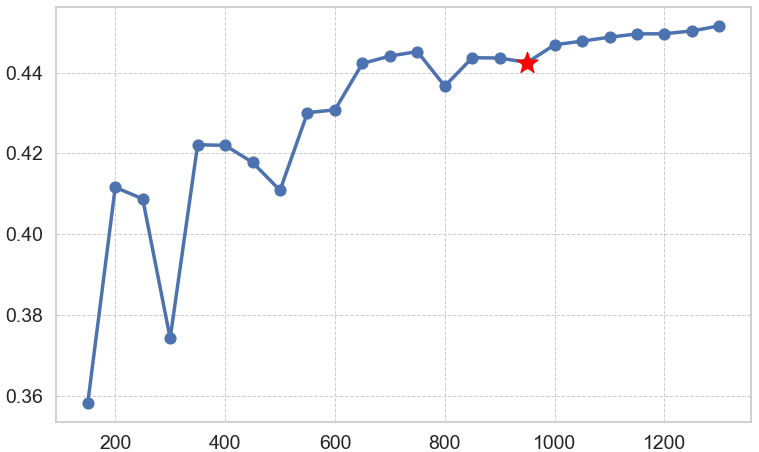}
    \caption{Training was conducted for 1300 epochs, with the performance metric IoU measured every 50 epochs, the loss function was introduced after epoch 950 (marked by a red pentagram). }
    \label{loss_constraint}
\end{figure}

To validate the effectiveness of the proposed loss function, we extended the number of training epochs and observed fluctuations in performance metrics during the bottleneck phase. After introducing our loss function at epoch 950, the performance further improved compared to previous stages, as illustrated in Fig.~\ref{loss_constraint}. This improvement is primarily attributed to the incorporation of adaptive weighting in sparse regions and geometric consistency constraints, which enable the model to more effectively capture key structural information and reduce prediction errors in sparse areas. 

\subsection{Failure Case}

\begin{figure}[htbp]
    \centering
    % 第一行左图
    \begin{minipage}[b]{0.48\columnwidth}
        \centering
        \begin{minipage}[c]{\linewidth}
            \centering
            \includegraphics[width=\linewidth]{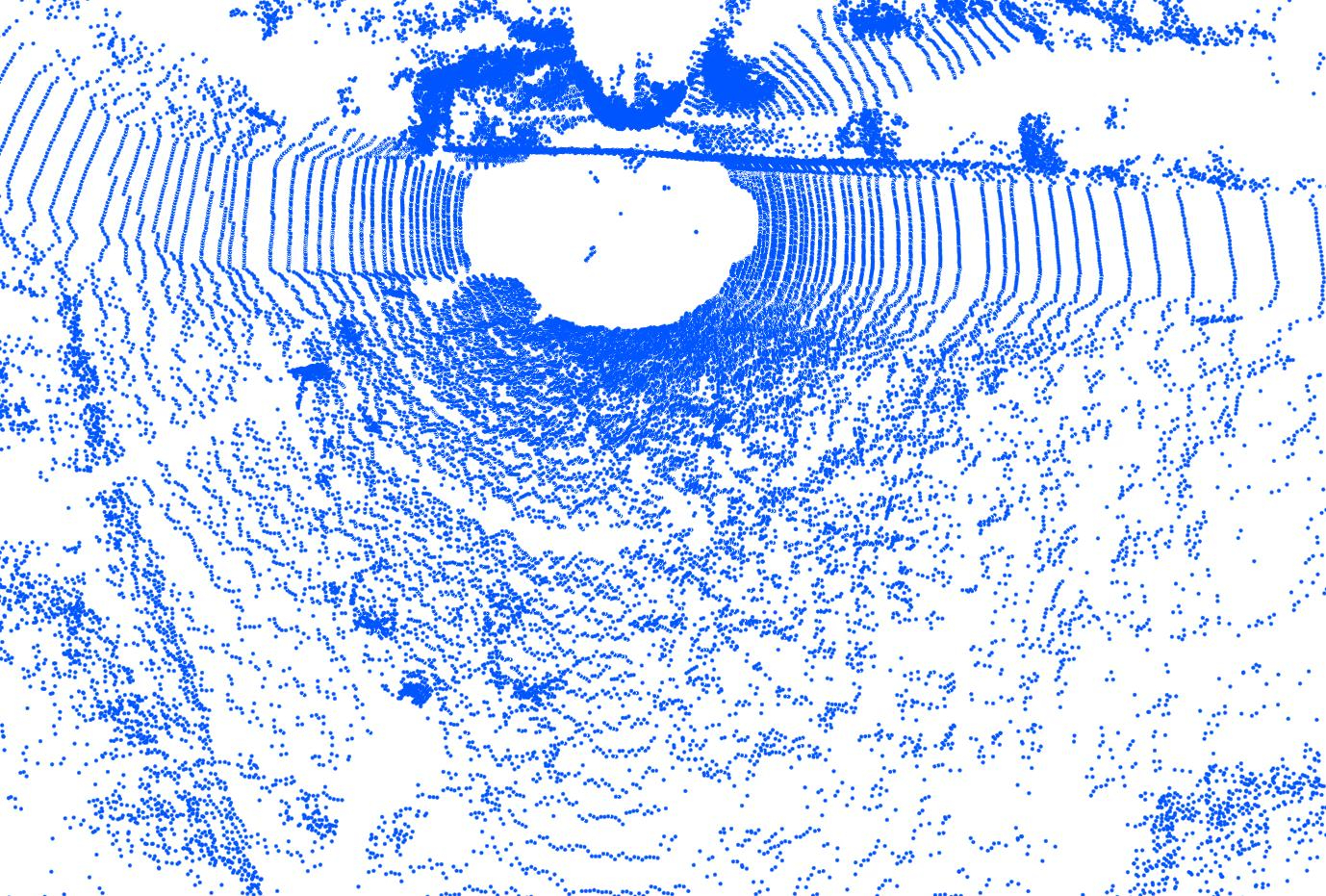} \\
            \scriptsize (a) Ground Truth
        \end{minipage}
    \end{minipage}%
    \hfill
    % 第一行右图
    \begin{minipage}[b]{0.48\columnwidth}
        \centering
        \begin{minipage}[c]{\linewidth}
            \centering
            \includegraphics[width=\linewidth]{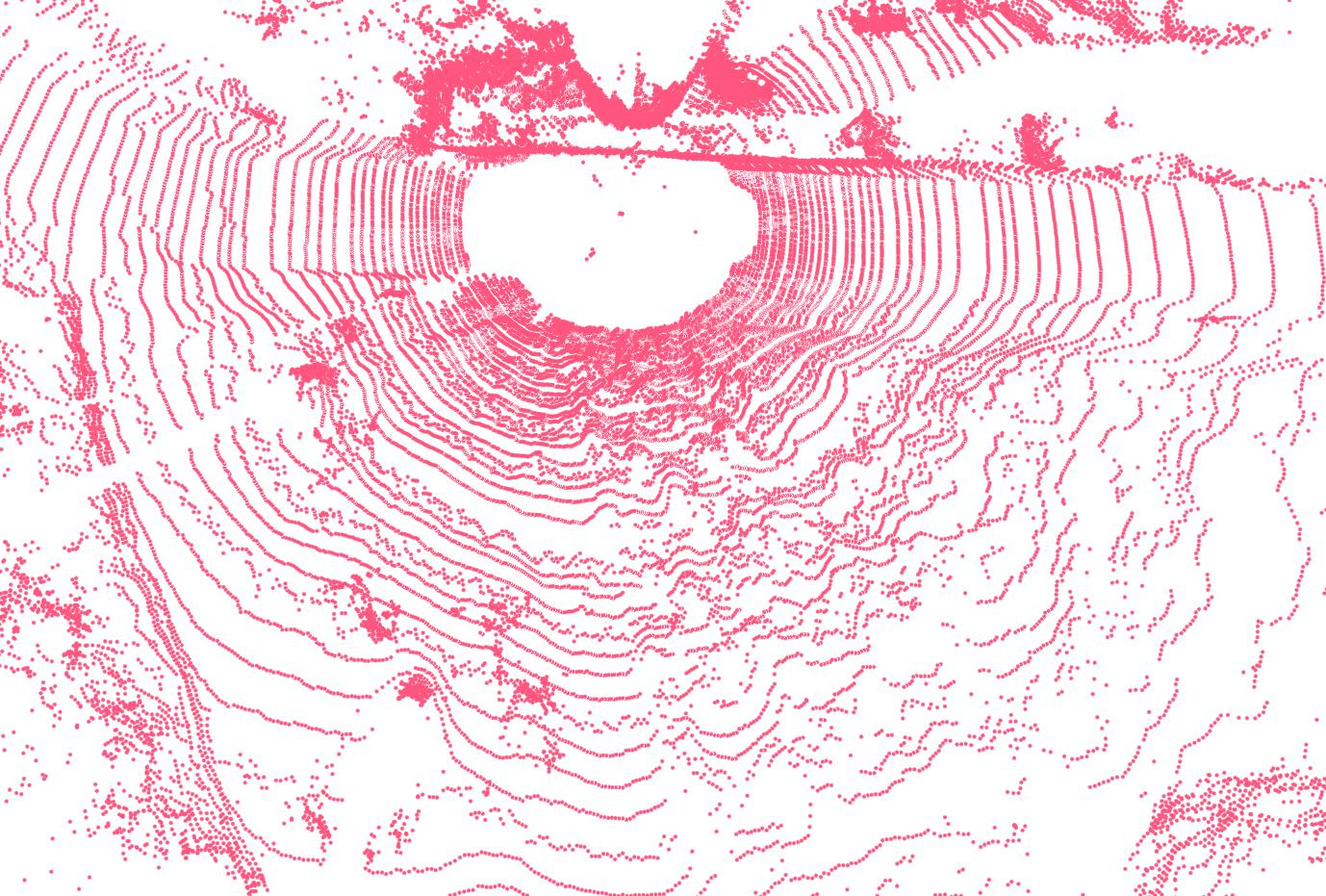} \\
            \scriptsize (b) Ours
        \end{minipage}
    \end{minipage}
    \caption{Failure case in a noisy scene with irregular vegetation. (a) Ground Truth. (b) Our. Erroneously reconstructs discrete points as continuous contour lines.}
    \label{Failure Case}
\end{figure}

Although our proposed method significantly outperforms other state-of-the-art approaches overall, it still exhibits certain limitations in upsampling quality under specific conditions due to the lack of spatial structural information in range image. For example, as shown in Fig.~\ref{Failure Case}, our method demonstrates inferior upsampling performance in certain noisy scenes. The irregular vegetation structure increases uncertainty during the reconstruction process. As observed in the details of Fig.~\ref{Failure Case}(b), the grass, which should be discretely distributed, is mistakenly interpreted as a continuous ground contour.

\section{Conclusion}
This paper proposes a super-resolution reconstruction method for automotive LiDAR point clouds, aiming to effectively suppress geometric distortion caused by ghost point noise. The method innovatively introduces a "scan-to-focus" driver visual mechanism that enhances global feature perception in range images by simulating the visual attention shift characteristics of human drivers during driving, generating high-resolution range images with improved geometric fidelity. Through learnable parameters, it adaptively focuses on sparse region features, significantly enhancing the upsampling accuracy of long-range point clouds. To address the inherent sparsity and discreteness of LiDAR point clouds, we propose incorporating a bird's-eye-view geometric constraint loss function, which effectively reduces the generation of spatially discrete noise points through multi-perspective supervision.

\appendix
\subsection{Appendx A}
Given a point $P=(x_{i},y_{i},z_{i})$, the spherical projection formulas are as follows:
\begin{equation}
    \left\{\begin{matrix}
    {v}_{i}= argmin(|\varphi_{b}-\arctan(\Delta_{b}-z_{i},\sqrt{x_{i}^{2}+y_{i}^{2}})|)
    \\
    u_i = \left(1 - (\arctan(y_i, x_i) + \pi)(2\pi)^{-1} \right) W
\end{matrix}\right.
\label{Hough Voting}
\end{equation}

where $\varphi_{b}$ and $\Delta_{b}$ denote arrays associated with the LiDAR beam, derived via a Hough voting mechanism to mitigate truncation errors in the projection process and produce higher-quality range images.

% \section{Reference Examples}

% \bibliographystyle{elsarticle-num-names} 
\bibliographystyle{ieeetr}
\bibliography{sample}

\begin{thebibliography}{10}

\bibitem{9969624}
K.~Matsuzaki and S.~Komorita, ``Efficient deep super-resolution of voxelized point cloud in geometry compression,'' {\em IEEE Sensors Journal}, vol.~23, no.~2, pp.~1328--1342, 2023.

\bibitem{s25092697}
C.~Chen, L.~Zhao, W.~Guo, X.~Yuan, S.~Tan, J.~Hu, Z.~Yang, S.~Wang, and W.~Ge, ``Farvnet: A fast and accurate range-view-based method for semantic segmentation of point clouds,'' {\em Sensors}, vol.~25, no.~9, 2025.

\bibitem{10657540}
S.~Zhao, Y.~Gao, T.~Wu, D.~Singh, R.~Jiang, H.~Sun, M.~Sarawata, Y.~Qiu, W.~Whittaker, I.~Higgins, Y.~Du, S.~Su, C.~Xu, J.~Keller, J.~Karhade, L.~Nogueira, S.~Saha, J.~Zhang, W.~Wang, C.~Wang, and S.~Scherer, ``Subt-mrs dataset: Pushing slam towards all-weather environments,'' in {\em 2024 IEEE/CVF Conference on Computer Vision and Pattern Recognition (CVPR)}, pp.~22647--22657, 2024.

\bibitem{11030950}
X.~Zeng, Y.~Hu, X.~Yang, Z.~Yin, and S.~Zhong, ``A multi-source fusion system for through-wall radar compensation using lidar and slam-based 3d reconstruction,'' {\em IEEE Sensors Journal}, pp.~1--1, 2025.

\bibitem{10657437}
B.~Yang, P.~Pfreundschuh, R.~Siegwart, M.~Hutter, P.~Moghadam, and V.~Patil, ``Tulip: Transformer for upsampling of lidar point clouds,'' in {\em 2024 IEEE/CVF Conference on Computer Vision and Pattern Recognition (CVPR)}, pp.~15354--15364, 2024.

\bibitem{chen2025srmambamambasuperresolutionlidar}
C.~Chen and W.~Ge, ``Srmamba: Mamba for super-resolution of lidar point clouds,'' 2025.

\bibitem{10.1007/978-3-031-72784-9_7}
Q.~Hu, Z.~Zhang, and W.~Hu, ``Rangeldm: Fast realistic lidar point cloud generation,'' in {\em Computer Vision -- ECCV 2024} (A.~Leonardis, E.~Ricci, S.~Roth, O.~Russakovsky, T.~Sattler, and G.~Varol, eds.), (Cham), pp.~115--135, Springer Nature Switzerland, 2025.

\bibitem{SHAN2020103647}
T.~Shan, J.~Wang, F.~Chen, P.~Szenher, and B.~Englot, ``Simulation-based lidar super-resolution for ground vehicles,'' {\em Robotics and Autonomous Systems}, vol.~134, p.~103647, 2020.

\bibitem{FU2025112132}
B.~Fu, M.~Sui, H.~Li, F.~Yang, H.~Yao, T.~Deng, and X.~Zhang, ``Ligapu: A lidar point cloud upsampling network for multiple complex scenes,'' {\em Pattern Recognition}, p.~112132, 2025.

\bibitem{Saalmann_Pigarev_Vidyasagar_2007}
Y.~B. Saalmann, I.~N. Pigarev, and T.~R. Vidyasagar, ``Neural mechanisms of visual attention: How top-down feedback highlights relevant locations,'' {\em Science}, p.~1612–1615, Jun 2007.

\bibitem{Gilbert_Sigman_2007}
C.~D. Gilbert and M.~Sigman, ``Brain states: Top-down influences in sensory processing,'' {\em Neuron}, vol.~54, p.~677–696, Jun 2007.

\bibitem{10.1007/978-3-031-72973-7_21}
M.~Zheng, L.~Sun, J.~Dong, and J.~Pan, ``Smfanet: A lightweight self-modulation feature aggregation network for efficient image super-resolution,'' in {\em Computer Vision -- ECCV 2024} (A.~Leonardis, E.~Ricci, S.~Roth, O.~Russakovsky, T.~Sattler, and G.~Varol, eds.), (Cham), pp.~359--375, Springer Nature Switzerland, 2025.

\bibitem{10.5555/3524938.3525585}
S.~Mittal, A.~Lamb, A.~Goyal, V.~Voleti, M.~Shanahan, G.~Lajoie, M.~Mozer, and Y.~Bengio, ``Learning to combine top-down and bottom-up signals in recurrent neural networks with attention over modules,'' in {\em Proceedings of the 37th International Conference on Machine Learning}, ICML'20, JMLR.org, 2020.

\bibitem{Chen_Sun_Tian_Shen_Huang_Yan_2020}
H.~Chen, K.~Sun, Z.~Tian, C.~Shen, Y.~Huang, and Y.~Yan, ``Blendmask: Top-down meets bottom-up for instance segmentation,'' in {\em 2020 IEEE/CVF Conference on Computer Vision and Pattern Recognition (CVPR)}, Jun 2020.

\bibitem{7780459}
K.~He, X.~Zhang, S.~Ren, and J.~Sun, ``Deep residual learning for image recognition,'' in {\em 2016 IEEE Conference on Computer Vision and Pattern Recognition (CVPR)}, pp.~770--778, 2016.

\bibitem{7478072}
E.~Shelhamer, J.~Long, and T.~Darrell, ``Fully convolutional networks for semantic segmentation,'' {\em IEEE Transactions on Pattern Analysis and Machine Intelligence}, vol.~39, no.~4, pp.~640--651, 2017.

\bibitem{Lou_2025_CVPR}
M.~Lou and Y.~Yu, ``Overlock: An overview-first-look-closely-next convnet with context-mixing dynamic kernels,'' in {\em Proceedings of the Computer Vision and Pattern Recognition Conference (CVPR)}, pp.~128--138, June 2025.

\bibitem{liu2024vmamba}
Y.~Liu, Y.~Tian, Y.~Zhao, H.~Yu, L.~Xie, Y.~Wang, Q.~Ye, J.~Jiao, and Y.~Liu, ``Vmamba: Visual state space model,'' {\em Curran Associates, Inc.}, vol.~37, pp.~103031--103063, 2024.

\bibitem{9710580}
Z.~Liu, Y.~Lin, Y.~Cao, H.~Hu, Y.~Wei, Z.~Zhang, S.~Lin, and B.~Guo, ``Swin transformer: Hierarchical vision transformer using shifted windows,'' in {\em 2021 IEEE/CVF International Conference on Computer Vision (ICCV)}, pp.~9992--10002, 2021.

\bibitem{964489}
M.~Alexa, J.~Behr, D.~Cohen-Or, S.~Fleishman, D.~Levin, and C.~Silva, ``Point set surfaces,'' in {\em Proceedings Visualization, 2001. VIS '01.}, pp.~21--29, 537, 2001.

\bibitem{10.1145/1073204.1073227}
S.~Fleishman, D.~Cohen-Or, and C.~T. Silva, ``Robust moving least-squares fitting with sharp features,'' {\em ACM Trans. Graph.}, vol.~24, p.~544–552, July 2005.

\bibitem{8099499}
R.~Q. Charles, H.~Su, M.~Kaichun, and L.~J. Guibas, ``Pointnet: Deep learning on point sets for 3d classification and segmentation,'' in {\em 2017 IEEE Conference on Computer Vision and Pattern Recognition (CVPR)}, pp.~77--85, 2017.

\bibitem{10.5555/3295222.3295263}
C.~R. Qi, L.~Yi, H.~Su, and L.~J. Guibas, ``Pointnet++: deep hierarchical feature learning on point sets in a metric space,'' in {\em Proceedings of the 31st International Conference on Neural Information Processing Systems}, NIPS'17, (Red Hook, NY, USA), p.~5105–5114, Curran Associates Inc., 2017.

\bibitem{9794916}
A.~Akhtar, Z.~Li, G.~V.~d. Auwera, L.~Li, and J.~Chen, ``Pu-dense: Sparse tensor-based point cloud geometry upsampling,'' {\em IEEE Transactions on Image Processing}, vol.~31, pp.~4133--4148, 2022.

\bibitem{10944132}
C.~Fan, C.~Zhao, and Y.~Duan, ``Pvt: An implicit surface reconstruction framework via point voxel geometric-aware transformer,'' in {\em 2025 IEEE/CVF Winter Conference on Applications of Computer Vision (WACV)}, pp.~3013--3023, 2025.

\bibitem{9969541}
D.~Guo, G.~Yang, B.~Qi, and C.~Wang, ``A fast ground segmentation method of lidar point cloud from coarse-to-fine,'' {\em IEEE Sensors Journal}, vol.~23, no.~2, pp.~1357--1367, 2023.

\bibitem{9612185}
Z.~Zhang, Z.~Liang, M.~Zhang, X.~Zhao, H.~Li, M.~Yang, W.~Tan, and S.~Pu, ``Rangelvdet: Boosting 3d object detection in lidar with range image and rgb image,'' {\em IEEE Sensors Journal}, vol.~22, no.~2, pp.~1391--1403, 2022.

\bibitem{10800127}
L.~Li, Q.~Sun, L.~Zhao, H.~Sun, F.~Zhao, and B.~Gu, ``Face mamba: A facial emotion analysis network based on vmamba*,'' in {\em 2024 7th International Conference on Machine Learning and Natural Language Processing (MLNLP)}, pp.~1--5, 2024.

\bibitem{10836868}
M.~Han, T.~Xu, Q.~Liu, X.~Yang, J.~Wang, and J.~Kong, ``Hfifnet: Hierarchical feature interaction network with multiscale fusion for change detection,'' {\em IEEE Journal of Selected Topics in Applied Earth Observations and Remote Sensing}, vol.~18, pp.~4318--4330, 2025.

\bibitem{10962143}
S.~Liu, Z.~Yang, Q.~Li, and Q.~Wang, ``Intermamba: A visual-prompted interactive framework for dense object detection and annotation,'' {\em IEEE Transactions on Geoscience and Remote Sensing}, vol.~63, pp.~1--11, 2025.

\bibitem{U-Net}
O.~Ronneberger, P.~Fischer, and T.~Brox, ``U-net: Convolutional networks for biomedical image segmentation,'' in {\em Medical Image Computing and Computer-Assisted Intervention -- MICCAI 2015} (N.~Navab, J.~Hornegger, W.~M. Wells, and A.~F. Frangi, eds.), (Cham), pp.~234--241, Springer International Publishing, 2015.

\bibitem{10958038}
Y.~Zhao, C.~Liu, X.~Zhou, and X.~Zhang, ``Segumamba: Integrating mamba with u\_net for medical image segmentation,'' in {\em 2024 International Conference on Image Processing, Computer Vision and Machine Learning (ICICML)}, pp.~108--111, 2024.

\bibitem{10.1007/978-981-97-5609-4_33}
X.~Yang, Z.~Luo, Y.~Wu, X.~Xie, L.~Nan, and T.~Li, ``Tmu: Transmission-enhanced mamba-unet for medical image segmentation,'' in {\em Advanced Intelligent Computing Technology and Applications} (D.-S. Huang, C.~Zhang, and J.~Guo, eds.), (Singapore), pp.~428--438, Springer Nature Singapore, 2024.

\bibitem{10.1007/978-981-96-2882-7_10}
W.~Hou, S.~Zhou, and H.~Zhao, ``Lightmamba-unet: Lightweight mamba with u-net for efficient skin lesion segmentation,'' in {\em Advances in Brain Inspired Cognitive Systems} (A.~Hussain, B.~Jiang, J.~Ren, M.~Mahmud, E.~Yang, A.~Zheng, C.~Li, S.~Wang, Z.~Gao, and Z.~Zhao, eds.), (Singapore), pp.~93--103, Springer Nature Singapore, 2025.

\bibitem{He_2025_CVPR}
H.~He, J.~Zhang, Y.~Cai, H.~Chen, X.~Hu, Z.~Gan, Y.~Wang, C.~Wang, Y.~Wu, and L.~Xie, ``Mobilemamba: Lightweight multi-receptive visual mamba network,'' in {\em Proceedings of the Computer Vision and Pattern Recognition Conference (CVPR)}, pp.~4497--4507, June 2025.

\bibitem{10.1007/978-3-031-91979-4_2}
T.~Huang, X.~Pei, S.~You, F.~Wang, C.~Qian, and C.~Xu, ``Localmamba: Visual state space model with windowed selective scan,'' in {\em Computer Vision -- ECCV 2024 Workshops} (A.~Del~Bue, C.~Canton, J.~Pont-Tuset, and T.~Tommasi, eds.), (Cham), pp.~12--22, Springer Nature Switzerland, 2025.

\bibitem{Hatamizadeh_2025_CVPR}
A.~Hatamizadeh and J.~Kautz, ``Mambavision: A hybrid mamba-transformer vision backbone,'' in {\em Proceedings of the Computer Vision and Pattern Recognition Conference (CVPR)}, pp.~25261--25270, June 2025.

\bibitem{8099589}
T.-Y. Lin, P.~Dollár, R.~Girshick, K.~He, B.~Hariharan, and S.~Belongie, ``Feature pyramid networks for object detection,'' in {\em 2017 IEEE Conference on Computer Vision and Pattern Recognition (CVPR)}, pp.~936--944, 2017.

\bibitem{9711179}
W.~Wang, E.~Xie, X.~Li, D.-P. Fan, K.~Song, D.~Liang, T.~Lu, P.~Luo, and L.~Shao, ``Pyramid vision transformer: A versatile backbone for dense prediction without convolutions,'' in {\em 2021 IEEE/CVF International Conference on Computer Vision (ICCV)}, pp.~548--558, 2021.

\bibitem{dosovitskiy2020image}
A.~Dosovitskiy, L.~Beyer, A.~Kolesnikov, D.~Weissenborn, X.~Zhai, T.~Unterthiner, M.~Dehghani, M.~Minderer, G.~Heigold, S.~Gelly, J.~Uszkoreit, and N.~Houlsby, ``An image is worth 16x16 words: Transformers for image recognition at scale,'' in {\em International Conference on Learning Representations (ICLR)}, 2021.

\bibitem{10678536}
Y.~Wang, Y.~Li, G.~Wang, and X.~Liu, ``Multi-scale attention network for single image super-resolution,'' in {\em 2024 IEEE/CVF Conference on Computer Vision and Pattern Recognition Workshops (CVPRW)}, pp.~5950--5960, 2024.

\bibitem{Hu_Shen_Sun_2018}
J.~Hu, L.~Shen, and G.~Sun, ``Squeeze-and-excitation networks,'' in {\em 2018 IEEE/CVF Conference on Computer Vision and Pattern Recognition}, Jun 2018.

\bibitem{liao2022kitti}
Y.~Liao, J.~Xie, and A.~Geiger, ``Kitti-360: A novel dataset and benchmarks for urban scene understanding in 2d and 3d,'' {\em IEEE Transactions on Pattern Analysis and Machine Intelligence}, vol.~45, no.~3, pp.~3292--3310, 2023.

\bibitem{fong2022panoptic}
W.~K. Fong, R.~Mohan, J.~V. Hurtado, L.~Zhou, H.~Caesar, O.~Beijbom, and A.~Valada, ``Panoptic nuscenes: A large-scale benchmark for lidar panoptic segmentation and tracking,'' {\em IEEE Robotics and Automation Letters}, vol.~7, no.~2, pp.~3795--3802, 2022.

\bibitem{zhang2024cas}
T.~Zhang, L.~Li, Y.~Zhou, W.~Liu, C.~Qian, and X.~Ji, ``Cas-vit: Convolutional additive self-attention vision transformers for efficient mobile applications,'' {\em arXiv preprint arXiv:2408.03703}, 2024.

\bibitem{9607618}
J.~Liang, J.~Cao, G.~Sun, K.~Zhang, L.~Van~Gool, and R.~Timofte, ``Swinir: Image restoration using swin transformer,'' in {\em 2021 IEEE/CVF International Conference on Computer Vision Workshops (ICCVW)}, pp.~1833--1844, 2021.

\end{thebibliography}

\end{document}